\documentclass[10pt,twocolumn,journal]{IEEEtran}

\usepackage{epsfig}
\usepackage{epstopdf}
\usepackage{citesort}
\usepackage{amsmath}
\usepackage{amssymb}
\usepackage{color}
\usepackage{adjustbox}
\usepackage{array}
\usepackage{multirow}
\usepackage{algorithm}
\usepackage{algorithmic}
\usepackage{rotating}
\usepackage{graphicx}
\usepackage{subfigure}
\usepackage[colorlinks,linkcolor=blue]{hyperref}
\usepackage{colortbl}
\usepackage{xcolor}
\definecolor{gray}{rgb}{0.91,0.91,0.91}
\usepackage{changepage}
\usepackage{fancyhdr}
\usepackage{booktabs}    
\usepackage{makecell}  
\usepackage{amsmath,amssymb}  %
\newlength{\figurewidth}
\newlength{\smallfigurewidth}

\begin{document}

\title
{
UniTS: Unified Spatio-Temporal Generative Model for Remote Sensing
}

\author{%
Yuxiang Zhang,
Shunlin Liang,~\IEEEmembership{Fellow,~IEEE},
Wenyuan Li,
Han Ma,
Jianglei Xu,
Yichuan Ma, \\
Jiangwei Xie,
Wei Li,~\IEEEmembership{Senior Member,~IEEE},
Mengmeng Zhang,~\IEEEmembership{Member,~IEEE}, \\
Ran Tao,~\IEEEmembership{Senior Member,~IEEE},
Xiang-Gen Xia,~\IEEEmembership{Fellow,~IEEE}\\
\thanks{%
 The research work was conducted in the JC STEM Lab of Quantitative Remote sensing funded by The Hong Kong Jockey Club Charities Trust. This work was supported in part by the Funds of the National Natural Science Foundation of China under Grant W2411055. (Corresponding author: Shunlin Liang; shunlin@hku.hk).  }
\thanks{%
	Y. Zhang, S. Liang, W. y. Li, H. Ma, J. Xu and Y. Ma are with \href{jcqrs.hku.hk}{the Jockey Club STEM Laboratory of Quantitative Remote Sensing}, Department of Geography, the University of Hong Kong, Hong Kong, China (e-mail: yxzhang7@hku.hk, shunlin@hku.hk, liwayne@hku.hk, mahan@hku.hk, jlxurs@hku.hk and yichuanm@hku.hk).
}
\thanks{%
	J. Xie, W. Li, M. Zhang and R. Tao are with the School of Information and Electronics, Beijing Institute of Technology, and Beijing Key Laboratory of Fractional Signals and Systems, 100081 Beijing, China (e-mail: xiejiangweiouc@gmail.com, liwei089@ieee.org, mengmengzhang@bit.edu.cn, rantao@bit.edu.cn).
}
\thanks{%
	X. Xia is with the Department of Electrical and Computer Engineering, University of Delaware, Newark, DE 19716, USA (e-mail: xianggen@udel.edu).
}
}

\maketitle
\fancyfoot[C]{\thepage}
\pagenumbering{arabic}
\renewcommand{\headrulewidth}{0pt}

\begin{abstract}
One of the primary objectives of satellite remote sensing is to capture the complex dynamics of the Earth environment, which encompasses tasks such as reconstructing continuous cloud-free image sequences, detecting land cover changes, and forecasting future surface evolution. However, existing methods typically require specialized models tailored to different tasks, and lack a general framework that can address these multi-level tasks from a unified perspective. In this paper, we propose a Unified Spatio-Temporal Generative Model (UniTS), which integrates several long-separated core tasks, including time series reconstruction, time series cloud removal, time series semantic change detection, and time series forecasting. Based on the flow matching generative paradigm, UniTS constructs a deterministic evolution path from noise to targets under the guidance of task-specific conditions, achieving unified modeling of spatiotemporal representations for multi-level tasks. The UniTS architecture  consists of a diffusion transformer with spatiotemporal blocks, where we design an Adaptive Condition Injector (ACor) to enhance the model's conditional perception of multimodal inputs, enabling high-quality controllable generation. Additionally, we design a Spatiotemporal-aware Modulator (STM) to improve the ability of spatiotemporal blocks to capture complex spatiotemporal dependencies. Furthermore, we construct two high-quality multimodal time series datasets, TS-S12 and TS-S12CR, filling the gap of benchmark datasets for time series cloud removal and forecasting tasks. Extensive experiments demonstrate that UniTS exhibits exceptional generative and cognitive capabilities across reconstruction, interpretation, and forecasting tasks. It substantially outperforms existing specialized models, particularly under challenging conditions such as severe cloud contamination, modality absence, and forecasting complex phenological variations. More details can be found on the project page: \href{https://yuxiangzhang-bit.github.io/UniTS-website/}{https://yuxiangzhang-bit.github.io/UniTS-website/}.
\end{abstract}

\begin{keywords}
Image sequences,
Time series reconstruction, 
Time series cloud removal, 
Time series semantic change detection, 
Time series forecasting,
Flow matching,
Generative model
\end{keywords}

\section{Introduction}
Satellite image time series have become an indispensable tool for understanding and monitoring the dynamics of Earth's systems, as they provide continuous spatiotemporal observational data. These long-term and consistent records of land surface dynamics deliver critical support for numerous fields, including ecological environment assessment, climate mitigation, and emergency response to natural disasters. With the increasing availability of high spatiotemporal resolution satellite data, time series analysis has expanded beyond single-sensor or single-modal approaches to integrate multispectral, radar, panchromatic, and other multi-source data, enabling more comprehensive and accurate perception of land surface changes.

\begin{figure*}[tp] \small
	\vspace{-4em}
	\begin{center}
		\epsfig{width=2.2\figurewidth,file=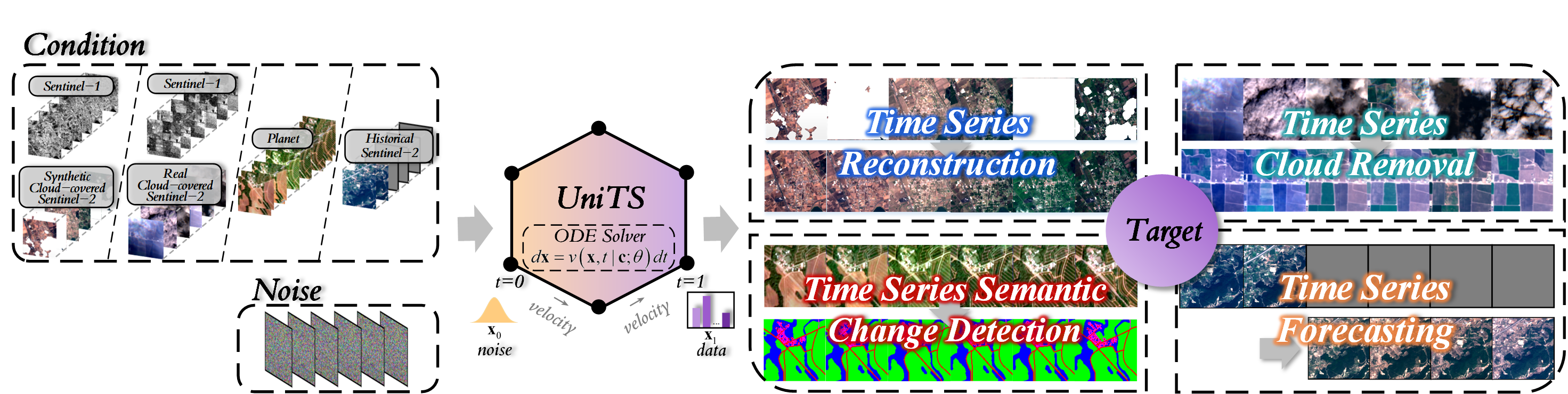}
	\end{center}
	\caption{\label{fig:UniTS-fm} The framework of UniTS applied to four remote sensing time series tasks. UniTS takes the conditions and noise from different time series tasks as input, and through a trained complete velocity field, gradually samples from the noise distribution to the target data distribution, producing time series outputs including cloud-free time series, semantic change maps, future predictions, and more.}
\end{figure*}

Based on the objectives and processing levels of time series analysis tasks, we categorize them into two broad classes: low-level and high-level vision tasks. \textbf{Low-level tasks} focus on improving data quality and completeness, primarily including \textbf{time series reconstruction} and \textbf{time series cloud removal}. The core objective is to recover missing pixels, suppress noise and cloud, and generate high-quality, contamination-free time series images. \textbf{High-level tasks} aim to extract high-level semantic information from the reconstructed clear data to achieve scene understanding and trend inference, mainly encompassing \textbf{time series semantic change detection} and \textbf{time series forecasting}.


Low-level tasks such as gap filling and cloud removal are essential in optical remote sensing due to frequent atmospheric interference. Clouds, haze, and cloud shadows often block or distort optical signals, causing missing or corrupted data. \emph{According to statistics from long-term observations by the MODIS sensor \cite{6422379}, approximately 67\% of the Earth's surface and 55\% of the land area are persistently affected by cloud cover on average.} This high coverage rate not only significantly reduces the practical observation efficiency of optical satellites but also makes completely cloud-free time series images extremely scarce at regional scales. Additional issues like sensor failures or transmission errors can further degrade data quality. The primary goal of time series reconstruction is to recover complete, high-quality image time series from incomplete or noisy observational data. Common approaches use cloud masks to simulate data gaps on clean sequences. Stucker et al. \cite{10319805} introduced U-TILISE, a representation learning framework using an encoder-decoder with temporal attention. From a generation viewpoint, Shu et al. \cite{shu2025restore} developed RESTORE-Diffusion Transformer (DiT), a multimodal diffusion framework that fuses optical and Synthetic Aperture Radar (SAR) time series with date embeddings to guide reconstruction under clouds.

Time series cloud removal is closely related to time series reconstruction and poses greater challenges. Unlike simply reconstructing areas obscured by cloud masks, this task specifically aims at restoring pixels occluded by clouds and their shadows within a temporal sequence. Patrick et al. \cite{ebel2022sen12ms} constructed the SEN12MS-CR-TS dataset, providing 30 temporally aligned Sentinel-1/2 image pairs with corresponding cloud masks. To address data gaps in visible and near-infrared bands, Gonzalez-Calabuig et al. \cite{gonzalez2025generative} proposed GANFilling, a GAN-based method that integrates convolutional LSTM layers to convert cloudy optical sequences into cloud-free ones.


The continuous spatiotemporal time series data used for understanding and analysis in high-level tasks is provided by low-level tasks. Time series semantic change detection monitors complex land cover dynamics by identifying not only the precise location (Where) and timing (When) of changes but also provide semantic information about the type of change (What). Garnot et al. \cite{garnot2021panoptic} proposed a U-Net with Temporal Attention Encoder (U-TAE) for time series panoptic segmentation. However, this method is only designed to predict a single land cover map corresponding to the time series images and does not support land cover change detection. He et al. \cite{he2024time} introduced an end-to-end framework based on a fully convolutional network, which directly learns the mapping between spectral features and land cover classes to achieve time series semantic change detection in a unified manner.

Time series forecasting predicts future surface conditions by analyzing historical Earth observation data and related factors, requiring models to capture temporal evolution patterns, seasonal trends, and long-term spatiotemporal dependencies. One early benchmark is the EarthNet2021 dataset by Requena‑Mesa et al. \cite{requena2021earthnet2021}, which pairs Sentinel‑2 imagery with topographic and meteorological variables for surface forecasting. Currently, Benson et al. \cite{10656834} introduced the GreenEarthNet dataset for high-resolution vegetation forecasting and developed a multimodal transformer model named Contextformer.


Satellite image time series hold significant value in remote sensing applications. With the increasing abundance of remote sensing data resources, we are now presented with a critical opportunity to construct a unified time series framework that can collaboratively serve multiple remote sensing time series tasks. To this end, our investigation into existing low-level and high-level time series tasks has revealed the following limitations:
\begin{itemize}
	
	\item \textbf{Limited exploration in time series cloud removal tasks and challenges in dataset construction.} Most studies focus on reconstructing time series using cloud masks, which oversimplify real-world conditions with variable cloud contamination. Progress in time series cloud removal is limited, and the currently available benchmark datasets exhibit significant shortcomings. For example, the SEN12MS-CR-TS dataset \cite{ebel2022sen12ms} suffers from temporal misalignment of up to 14 days and excludes images with more than 50\% cloud cover, while the EarthNet2021 dataset \cite{requena2021earthnet2021} also omits heavily clouded sequences. Consequently, the absence of high-quality, temporally aligned time series datasets severely hinders reliable model training and evaluation in this field.
	
	
	\item \textbf{Limited exploration in time series forecasting tasks.} Most existing studies employ discriminative models, such as ConvLSTM and 3D CNN, exemplified by Pangu-Weather \cite{bi2023accurate}. However, these models struggle to capture the complexities of spatiotemporal data. There is a need to explore generative models in time series forecasting research that can better fit complex distributions. Current studies mainly focus on predicting vegetation indices, with limited work on forecasting original remote sensing images, especially for high-resolution multispectral data like Sentinel-2. Due to its high spectral dimensionality and complex spatial details, predicting original multispectral time series presents even greater challenges.

	\item \textbf{Lack of a unified framework for multiple remote sensing time series tasks.} Current research remains largely in the stage of developing specialized models tailored for specific tasks, and a unified framework that can effectively address diverse time series tasks is still missing. We argue that the core challenge for both low-level and high-level time series tasks lies in mining spatiotemporal representations, with the main difference being how these representations are adapted through task-specific designs to meet particular requirements.
	
\end{itemize}

To address the aforementioned technical and data gaps, we propose Unified Spatio-Temporal Generative Model (UniTS) based on the flow matching-based generative paradigm, which is applicable to four time series tasks, as shown in Fig.\ref{fig:UniTS-fm}. Additionally, we construct two high-quality benchmarks TS-S12 and TS-S12CR for time series model validation. The main contributions of this paper are as follows:

\begin{itemize}
	
	\item UniTS achieves a unified modeling for the first time in multi-level tasks: time series reconstruction, time series cloud removal, time series semantic change detection, and time series forecasting. The framework demonstrates excellent generation and understanding capabilities across different task levels, whether it is low-level time series reflectance recovery or high-level time series semantic understanding.
	\item Within the diffusion transformer framework, we introduce a spatiotemporal block and design two novel components: the Adaptive Condition Injector (ACor) and the Spatiotemporal-aware Modulator (STM). ACor adaptively injects multimodal conditional information (e.g., SAR and optical imagery) by dynamically generating affine transformation parameters, significantly enhancing the model's conditional perception of multimodal inputs across various time series tasks. Meanwhile, STM modulates attention weights in the spatiotemporal block by leveraging generated dynamic bias terms based on spatiotemporal priors, thereby strengthening the model’s capacity to capture complex spatiotemporal dependencies.	
	\item We construct two high-quality multimodal time series datasets, namely TS-S12 and TS-S12CR. Among them, TS-S12 and TS-S12CR contain Sentinel-1 and Sentinel-2 imagery from 14,973 and 12,126 ROIs around the world, respectively. TS-S12 provides aligned sample pairs of Sentinel-1 and cloud-free Sentinel-2 for time series reconstruction and forecasting tasks. TS-S12CR offers aligned triplet samples of Sentinel-1, cloud-covered Sentinel-2, and cloud-free Sentinel-2 specifically designed for time series cloud removal task. TS-S12CR provides an extreme scenario with an average cloud coverage of 84.02\%, serving as an important benchmark for developing robust time series cloud removal methods.
	
\end{itemize}


The rest of the paper is organized as follows. Section \ref{sec:Datasets} describes the two time series benchmarks. Section \ref{sec:proposed} presents the details of the proposed UniTS. The extensive experiments and analyses on four time series tasks are presented in Section \ref{sec:results}. Finally, conclusions are drawn in Section \ref{sec:conclusions}.


\begin{figure*}[tp] \small
	\vspace{-4em}
	\begin{center}
		\epsfig{width=2.2\figurewidth,file=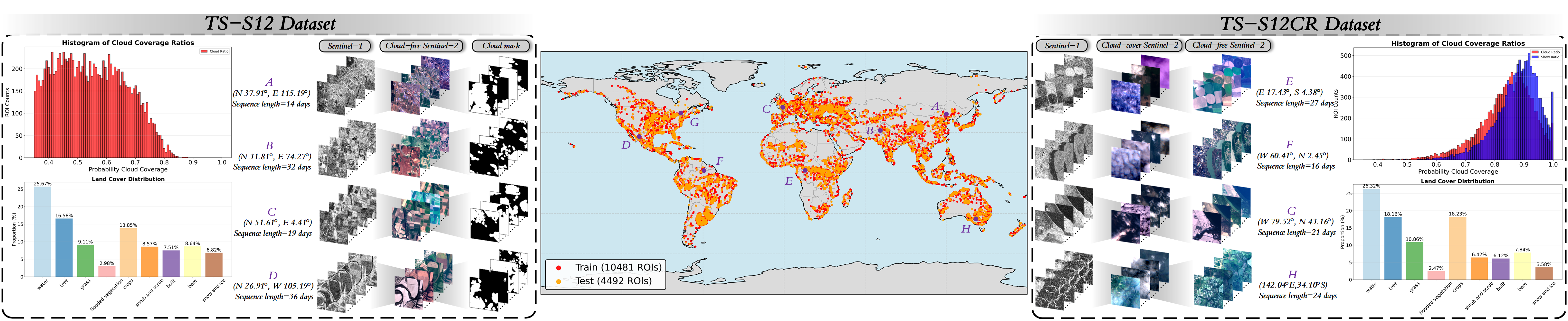}
	\end{center}
	\caption{\label{fig:dataset} Left: TS-S12 dataset; middle: Geographical distribution of ROIs; right: TS-S12CR dataset.}
\end{figure*}

\begin{table*}[]
	\setlength\tabcolsep{3pt}
	\caption{Summary of TS-S12 and TS-S12CR datasets.}
	\label{tab:dataset}
	\centering
	\begin{tabular}{c|ccccccc}
		\toprule
		Dataset &
		Task &
		ROIs &
		Satellites &
		Time series &
		Images &
		Training/test &
		Cloud/shadow coverage \\ \midrule
		TS-S12 &
		\begin{tabular}[c]{c}Time Series Reconstruction \\ Time Series Forecasting\end{tabular}  &
		14,973 &
		Sentinel-1\&2 &
		8$\sim$97 (med-21) &
		740,094 &
		10,481/4,492 &
		Cloud: 2\%$\sim$88\% (mean-48.81\%) \\
		TS-S12CR &
		Time Series Cloud Removal &
		12,126 &
		Sentinel-1\&2 &
		8$\sim$44 (med-17) &
		626,751 &
		8,488/3,638 &
		\begin{tabular}[c]{c}Cloud: 30\%$\sim$100\% (mean-84.02\%)\\ Shadow: 38\%$\sim$100\% (mean-87.54\%)\end{tabular}
		\\ \bottomrule
	\end{tabular}
\end{table*}


\section{Datasets}
\label{sec:Datasets}
To address the current lack of high-quality paired multi-modal time series data, particularly for time series cloud removal tasks, we construct the TS-S12 and TS-S12CR datasets based on AllClear \cite{zhou2024allclear}.

\subsection{Data Preparation}

We select tens of thousands of ROIs worldwide, utilizing multispectral optical images from Sentinel-2A/B and SAR images from Sentinel-1A/B, all acquired in 2022. For Sentinel-2 data, we employ Level-1C orthorectified Top-of-Atmosphere (TOA) reflectance products, retaining all 10 spectral bands (excluding B1 Aerosols, B9 Water Vapor, and B10 Cirrus). For Sentinel-1 data, we use the Ground Range Detected (GRD) product, which includes dual-polarization channels (VV and VH). Additionally, we collect the corresponding Dynamic World Land Cover Map \cite{brown2022dynamic} for all Sentinel-2 images across the ROIs. Each ROI corresponds to a 2.56$\times$2.56 km² (256$\times$256) patch with a spatial resolution of 10m. Cloud masks are derived from the cloud probability dataset generated by the S2cloudless algorithm, while shadow masks are adopted from those produced in AllClear.

\subsection{TS-S12 Dataset}

The TS-S12 dataset contains 14,973 ROIs distributed across the globe, covering diverse land cover types. For each ROI, we construct aligned sample pairs consisting of Sentinel-1 data and cloud-free Sentinel-2 data. The sample selection and alignment strategies are as follows:

\begin{itemize}
	
	\item Cloud filtering: Cloud-free Sentinel-2 samples with cloud coverage less than 15\% and shadow coverage less than 30\% are selected.
	\item Temporal window: Sentinel-1 images captured within three days before and after the acquisition time of Sentinel-2 are matched, with the acquisition time of Sentinel-2 as the reference.
	
\end{itemize}

Using the above strategies, we filter the full-year 2022 data for each ROI using cloud and shadow masks, obtaining multi-modal time series data with sequence lengths ranging from 8$\sim$97. Furthermore, to support time series reconstruction and forecasting tasks, we retain all annual cloud masks for each ROI to simulate scenarios of missing data or cloud cover. Fig.\ref{fig:dataset} displays the histogram of cloud coverage ratios, land cover distribution, and examples of sample pairs from the TS-S12 dataset, with additional details provided in Table~\ref{tab:dataset}.

\subsection{TS-S12CR Dataset}

The TS-S12CR dataset comprises 12,126 ROIs distributed across the globe. We construct aligned sample pairs consisting of Sentinel-1 data, cloud-covered Sentinel-2 data, and cloud-free Sentinel-2 data. The sample selection and alignment strategies are as follows:

\begin{itemize}
	
	\item Cloud filtering: Cloud-free Sentinel-2 samples with cloud coverage less than 15\% and shadow coverage less than 30\% are selected.
	\item Temporal window: We use the acquisition time of the cloud-free Sentinel-2 image as the reference to match Sentinel-1 and cloud-covered Sentinel-2 images captured within a three-day window (before and after).
	
\end{itemize}

Based on the above strategies, we filter the data using cloud and shadow masks to obtain paired time series data with sequence lengths ranging from 8$\sim$44, suitable for the time series cloud removal task. It is worth noting that the average cloud/shadow coverage in this dataset are 84.02\% and 87.54\%. The high cloud coverage poses a significant challenge, requiring time series models to reconstruct surface information under extreme occlusion conditions. This not only tests the model's spatial restoration capability but also places higher demands on its ability to understand temporal evolution patterns and integrate multi-source information. Further details are shown in Fig.\ref{fig:dataset} and Table~\ref{tab:dataset}.

\section{Unified Time Series Generative Model}
\label{sec:proposed}
\definecolor{myblue}{RGB}{46,117,182}
\definecolor{myred}{RGB}{192,0,0}
\definecolor{mypurple}{RGB}{135,98,232}

In this section, we first introduce the framework of UniTS in Section \ref {subsec:Framework}. The details of ACor and STM are presented in Section \ref {subsec:ACor} and \ref {subsec:STM}, respectively. Finally, we elaborate on the training and inference processes for conditional time series generation in Section \ref{subsec:Traintest}. Notations used in this paper are summarized in Table \ref{table:Notations}.

\begin{table}[]
	\caption{\label{table:Notations}
		Notations of variables.}
	\setlength\tabcolsep{1pt}
	\begin{center}
		\begin{tabular}{ll}
			\toprule
			Notations                               & Description                                                 \\ \hline
			${{\bf{X}}} = \left\{ {{\bf{x}}_t}\!:\!t\in\![0, 1]\right\}\subset\mathbb{R}{^{T\!\times\!C\!\times\!H\times\!W}}$       & \begin{tabular}[c]{l}The set of samples corresponding to \\ flow time $t$\end{tabular}            \\
			${{\bf{x}}_\text{con}} \in\mathbb{R}{^{T\!\times\!{C_\text{con}}\!\times\!H\!\times\!W}}$                              & The task-specific conditions                     \\
			${\bf{z}} = \left\{ {{{\bf{z}}_t},{{\bf{z}}_\text{con}}} \right\} \subset\mathbb{R}{^{T\!\times\!d\!\times\!{n_\text h}\!\times\!{n_\text w}}}$       & \begin{tabular}[c]{l}The time series tokens obtained by \\ patch embedding\end{tabular}\\
			$\left\{ {{\bf{z}}_t^\text s,{\bf{z}}_\text{con}^\text s} \right\} \subset\mathbb{R}{^{T\!\times\!{n_\text h}{n_\text w}\!\times\!d}}$       & Input of spatial block  \\
			$\left\{ {{\bf{z}}_t^\text t,{\bf{z}}_\text{con}^\text t} \right\} \subset\mathbb{R}{^{{n_\text h}{n_\text w}\!\times\!T\!\times\!d}}$ & Input of temporal block  \\
			${\bf{m}}_{\text{\tiny DOY}} \in\mathbb{R}{^T}$ \& ${\bf{z}}_{\text{\tiny DOY}} \in\mathbb{R}{^{T\!\times d}}$      & Date list \& Date embedding   \\
			${\bf{m}}_{\text{\tiny lonlat}} \in\mathbb{R}{^2}$ \& ${\bf{z}}_{\text{\tiny lonlat}} \in\mathbb{R}{^{d}}$     & lon.\&lat. \& Geographic embedding \\
			${\bf{z}}_{\scriptscriptstyle\text{FM}}^\text s \in\mathbb{R}{^{{n_\text h}{n_\text w}\!\times\!d}}$      & The spatial timestep embedding   \\
			${\bf{z}}_{\scriptscriptstyle\text{FM}}^\text t \in\mathbb{R}{^{T\!\times d}}$       & The temporal timestep embedding \\
			${\bf{z}}_{\scriptscriptstyle\text{FM}}^\text{con} \in\mathbb{R}{^{{T_{\scriptscriptstyle\text{his}}} \times d}}$       & \begin{tabular}[c]{l}The historical condition \\ timestep embedding\end{tabular}  \\ \bottomrule
			&                                                             \\
			&
		\end{tabular}
	\end{center}
\vspace{-4em}
\end{table}

\subsection{Overview}
\label{subsec:Overview}

UniTS is a conditional time series generation model built on the flow matching paradigm. As illustrated in Fig.\ref{fig:UniTS-fm}, its core idea is to adapt to multiple time series tasks through a unified generative architecture. To achieve this, we organize unimodal or multimodal information, such as Sentinel-1 \& synthetic cloud-covered Sentinel-2 data, historical Sentinel-2 sequence segments, etc., into conditional signals tailored to different tasks. At the input stage, the task-specific conditions are concatenated with a random noise vector sampled from a standard Gaussian distribution, forming the input to UniTS. During training, UniTS employs an Ordinary Differential Equation (ODE) solver to learn a velocity field, which defines a deterministic path/flow from a simple noise distribution to the complex distribution of real time series data. The flow matching process constructs a continuous and smooth transformation trajectory, ensuring stability and efficiency throughout the generation process. In the inference phase, given a task-specific condition, the model first samples a random noise vector. Guided by the condition, the trained UniTS model then performs multi-step sampling by solving the learned ODE, following the pre-defined deterministic path. The noise is progressively refined and transformed into high-quality target time series data that meet task requirements. This framework unifies conditional generation as a flow-based transformation from noise to data, combining strong generative performance with remarkable flexibility.

\subsection{Preliminary: Flow Matching}

Denoising Diffusion Probabilistic Models (DDPMs) \cite{ho2020denoising} and their counterparts \cite{xie2025progressive,wang2025videoscene,11123732} based on Stochastic Differential Equations (SDEs) \cite{songscore} have set a remarkable benchmark in generative modeling. However, the multi-step iterative denoising and stochastic sampling mechanism inherent in DDPMs introduce fundamental limitations. 
Recently, flow matching \cite{liu2023flow,lipman2023flow,albergo2023building} has emerged as a novel generative paradigm, demonstrating remarkable advantages. By redefining the generation process as a flow along a deterministic velocity field, this approach not only circumvents the efficiency bottleneck of multi-step stochastic sampling in diffusion models but also ensures determinism and controllability in the generation process.

Flow matching constructs a time-dependent probability density path \({p}(\mathbf{x}, t)\), defining a continuous time stochastic process ${\bf{x}}(t)$ from the prior (noise) distribution to the target data distribution, where flow time \(t\in[0, 1]\). Given that \({p}(\mathbf{x}, 0)={p_0}(\mathbf{x})\) and \({p}(\mathbf{x}, 1)={p_1}(\mathbf{x})\) represent the prior distribution (e.g., standard Gaussian distribution) and the target data distribution, respectively, the objective is to derive \({p}(\mathbf{x}, t)\). This path \({p}(\mathbf{x}, t)\) is defined by a time-dependent velocity field ${{\bf{v}}({\bf{x}}(t), t)}$. The evolution of \({{\bf{x}}}(t)\) follows an ODE defined by this velocity field,
\begin{equation}\label{eq:1}
\frac{{d{\bf{x}}}(t)}{{dt}} = {{\bf{v}}({\bf{x}}(t), t)}
\end{equation}
The flow matching objective is to minimize the following objective function,
\begin{equation}\label{eq:3}
{{\cal L}_\text{FM}}(\theta ) = {\mathbb{E}_{{{\bf{x}}(t)},t}}\left[ {{{\left\| {{f_\theta }({{\bf{x}}(t)},t) - {{\bf{v}}({\bf{x}}(t), t)}} \right\|}^2}} \right]
\end{equation}
where $\theta$ denotes the learnable parameters, ${f_\theta }(\cdot)$ represents a neural network used for predicting the velocity field, $t$ is treated as a random variable and follows the uniform distribution on the interval [0, 1], i.e., $t\!\sim\!{\cal U}\left[ {0,1} \right]$. Directly regressing the velocity field is intractable. The key innovation of flow matching lies in decomposing this problem into a simpler conditional problem. Given a noise sample ${{\bf{x}}_0}\!\sim\! \mathcal{N}(0,I)$ and a target sample ${{\bf{x}}_1}\!\sim {p_1}\!({\bf{x}})$, they are connected via a conditional probability path \( p({\bf{x}},t|{{\bf{x}}_1})\). We define the conditional path based on the optimal transport path as proposed in \cite{lipman2023flow},
\begin{equation}\label{eq:2}
{{\bf{x}}(t)} = (1 - t) \cdot {{\bf{x}}_0} + t \cdot {{\bf{x}}_1}
\end{equation}
The conditional velocity field \( {\bf{v}}({\bf{x}},t|{{\bf{x}}_1})\) that generates this conditional probability path can be derived by taking the time derivative of ${\bf{x}}(t)$, that is \(\mathbf{x}_1 - \mathbf{x}_0\). Eq.(\ref{eq:3}) is rewritten as follows,
\begin{equation}\label{eq:3.1}
{{\cal L}_\text{FM}}(\theta ) = {\mathbb{E}_{{{\bf{x}}_0},{{\bf{x}}_1},t}}\left[ {{{\left\| {{f_\theta }({{\bf{x}}(t)},t) - {\bf{v}}({\bf{x}},t|{{\bf{x}}_1})} \right\|}^2}} \right]
\end{equation}
where ${{\bf{x}}(t)}$ is computed via Eq.(\ref{eq:2}) and ${\bf{v}}({\bf{x}},t|{{\bf{x}}_1})=\mathbf{x}_1 - \mathbf{x}_0$. By fitting the linear paths \(\mathbf{x}_1 - \mathbf{x}_0\) constructed from sufficient noise samples and target samples, the model implicitly learns how to map the noise distribution to the target data distribution. Remarkably, \cite{lipman2023flow} proves that minimizing this simple regression loss Eq.(\ref{eq:3.1}) is equivalent to regressing the real and intractable velocity field. During the inference phase, sampling becomes a straightforward process of solving the learned ODE. 


\subsection{UniTS Framework}
\label{subsec:Framework}
Assume that ${{\bf{X}}} = \left\{ {{\bf{x}}_t}\!:\!t\in\![0, 1]\right\}\subset\mathbb{R}{^{T\!\times\!C\!\times\!H\times\!W}}$ is the set of samples for flow time $t\in [0,1]$, ${{\bf{x}}_0}$ and ${{\bf{x}}_1}$ denote noise and target time series clips, respectively, and ${{\bf{x}}_\text{con}} \in\mathbb{R}{^{T\!\times\!{C_\text{con}}\!\times\!H\!\times\!W}}$ is a task-specific condition. Here, $T, H, W$ represent the length, height and width of time series, $C$ denotes the channel dimension for noise and target time series, and $C_\text{con}$ represents the channel dimension for condition. ${{\bf{x}}_t}$ is a realization of stochastic process ${{\bf{x}}(t)}$.
\begin{figure*}[tp] \small
	\begin{adjustwidth}{-1.5cm}{-1.5cm}
	\vspace{-4em}
	\begin{center}
		\epsfig{width=2.3\figurewidth,file=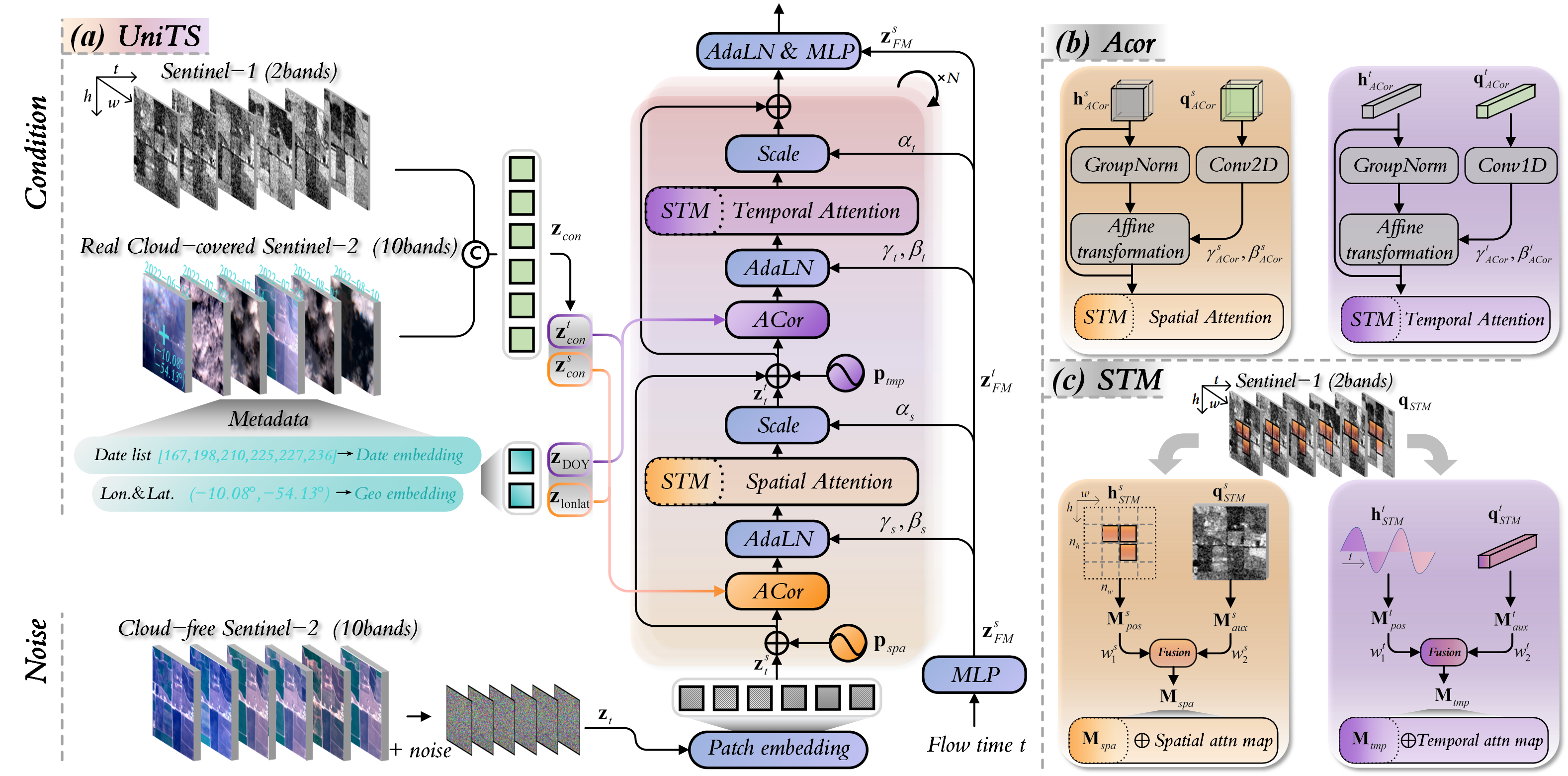}
	\end{center}
\end{adjustwidth}
	\caption{\label{fig:UniTS} The framework of UniTS. Taking the time series cloud removal task as an example. (a) UniTS architecture, (b) Adaptive Condition Injector (ACor), (c) Spatiotemporal-aware Modulator (STM).}
\end{figure*}

\begin{table*}[]
	\centering
	\setlength\tabcolsep{1pt}
	\caption{Summary of task-specific conditions, target samples, and metadata corresponding to all tasks. DOY represents Day of the Year, lon.\&lat. is longitude\&latitude.}
	\label{tab:input}
	\begin{adjustwidth}{-0.8cm}{-0.8cm}
		\begin{tabular}{c|cccc}
			\toprule
			Task                                                   & Dataset         & Task-specific conditions (${{\bf{x}}_{con}}$)        & Target samples (${{\bf{x}}_{1}}$)             & Metadata (${{\bf{m}}_{\text{\tiny DOY}},{\bf{m}}_{\text{\tiny lonlat}}}$)                       \\ \midrule
			Time Series Reconstruction               & TS-S12   & Sentinel-1 \& Synthetic cloud-covered Sentinel-2, \scriptsize{$C_\text{con}=12$} & Cloud-free Sentinel-2, \scriptsize{$C=10$}        & DOY, lon.\&lat.        \\  \midrule
			Time Series Cloud Removal                & TS-S12CR & Sentinel-1 \& Real cloud-covered Sentinel-2, \scriptsize{$C_\text{con}=12$}      & Cloud-free Sentinel-2, \scriptsize{$C=10$}       & DOY, lon.\&lat.        \\ \midrule
			\multirow{2}{*}{\begin{tabular}[c]{c}Time Series \\ Semantic Change Detection\end{tabular}} & DynamicEarthNet & Planet (RGBN), \scriptsize{$C_\text{con}=4$}                        & Segmentation map, \scriptsize{$C=6$}            & -                               \\
			& MUDS            & Planet (RGB), \scriptsize{$C_\text{con}=3$}                         & Segmentation map, \scriptsize{$C=2$}            & -                               \\ \midrule
			\multirow{2}{*}{Time Series Forecasting} & TS-S12   & Historical cloud-free Sentinel-2, \scriptsize{$C_\text{con}=10$}               & Future cloud-free Sentinel-2, \scriptsize{$C=10$} & Future DOY, lon.\&lat. \\
			& GreenEarthNet   & \begin{tabular}[c]{c}Historical Sentinel-2 (RGBN) \&\\ Meteorological observations \& Elevation map\end{tabular}, \scriptsize{$C_\text{con}=6$} & Future Sentinel-2 (RGBN), \scriptsize{$C=4$} & Future DOY, lon.\&lat. \\ \bottomrule
		\end{tabular}
	\end{adjustwidth}
\end{table*}

\textbf{Architecture.} We instantiate ${f_\theta}(\cdot)$ in Eq.(\ref{eq:3}) using a transformer-based architecture. This architecture is chosen for its simplicity, scalability, and effectiveness in generative modeling. Specifically, UniTS is implemented based on the standard DiT \cite{peebles2023scalable}, as shown in Fig.\ref{fig:UniTS}(a). We introduce interleaved spatial and temporal attention to construct a spatiotemporal block, in which two new components, ACor and STM, are designed to enhance time series conditional guidance and explore complex spatiotemporal correlations. Details are provided in Sections \ref{subsec:ACor} and \ref{subsec:STM}.

\textbf{Embedding of Task-specific Conditions and Metadata.} In UniTS, the task-specific conditions, target time series samples, and metadata corresponding to the four time series tasks are shown in Table~\ref{tab:input}. We introduce a patch embedding layer for the input intermediate state ${{\bf{x}}_t}$ at flow time $t$ and condition ${{\bf{x}}_\text{con}}$. Unlike previous latent transformers \cite{wang2025seedvr,zhang2024motiondiffuse,yan2025long} that operate on latents encoded by variational autoencoder (VAE), UniTS directly tokenizes raw pixel inputs. We adapt the patch embedding from Vision Transformer (ViT) to each image in the time series, obtaining tokens ${\bf{z}} = \left\{ {{{\bf{z}}_t},{{\bf{z}}_\text{con}}} \right\} \subset\mathbb{R}{^{T\!\times\!d\!\times\!{n_\text h}\!\times\!{n_\text w}}}$, where $d$ represents the channel dimension of each token, ${n_\text h}$ and ${n_\text w}$ are equal to $\frac{H}{h}$ and $\frac{W}{w}$, respectively. This process extracts non-overlapping image patches of size $h\times w$ from the time series for encoding. The tokens are then combined with 2D sincos positional encoding and fed into the spatiotemporal block.

Metadata includes the capture date list ${\bf{m}}_{\text{\tiny DOY}} \in\mathbb{R}{^T}$ (formatted as Day of the Year, DOY) for each image in the time series, along with the corresponding longitude and latitude coordinates ${\bf{m}}_{\text{\tiny lonlat}} \in\mathbb{R}{^2}$ for each sequence. We encode ${\bf{m}}_{\text{\tiny DOY}}$ using sinusoidal functions to obtain date embedding ${\bf{z}}_{\text{\tiny DOY}} \in\mathbb{R}{^{T\!\times d}}$, and jointly encode ${\bf{m}}_{\text{\tiny lonlat}}$ using a combination of Random Fourier Features and sinusoidal encoding to derive geographic embedding ${\bf{z}}_{\text{\tiny lonlat}} \in\mathbb{R}{^{d}}$. Through the above encoding scheme, we explicitly inject spatiotemporal priors into the model. By leveraging absolute spatiotemporal coordinates, the model can effectively reason under irregular observation intervals and generalize to unseen times and locations by capturing underlying periodic patterns and geospatial relationships, which is crucial for time series analysis \cite{garnot2020satellite}.

\textbf{Spatial and Temporal Positional Embeddings \& Flow Time Embedding.} Spatial and temporal positional embeddings are respectively added to the tokens in the first layer of the spatiotemporal block, enabling the model to understand the potential relationships of time series in both spatial and temporal dimensions. We introduce learnable spatial and temporal positional embeddings ${{\bf{p}}_{\text{spa}}} \in\mathbb{R}{^{{n_\text h}{n_\text w}\!\times d}},{{\bf{p}}_{\text{tmp}}} \in\mathbb{R}{^{T\!\times d}}$, which are initialized using 2D sincos positional encoding on a 2D grid of size ${n_\text h}\!\times\!{n_\text w}$, and 1D sincos positional encoding on a 1D time sequence of length $T$ respectively. After initialization, the spatial dimension of the spatial positional embedding is reshaped to ${n_\text h}{n_\text w}$. During training, a flow time $t$ is randomly sampled from the interval \([0, 1]\) as the timestep and then transformed into sinusoidal timestep embedding ${{\bf{z}}_{\scriptscriptstyle\text{FM}}} \in\mathbb{R}{^d}$. To enable the spatiotemporal block to construct an adaptive velocity field along the continuous path defined by the flow time, the timestep embedding ${{\bf{z}}_{\scriptscriptstyle\text{FM}}}$ is repeated in both spatial and temporal dimensions to obtain ${\bf{z}}_{\scriptscriptstyle\text{FM}}^\text s \in\mathbb{R}{^{{n_\text h}{n_\text w}\!\times\!d}}$ and ${\bf{z}}_{\scriptscriptstyle\text{FM}}^\text t \in\mathbb{R}{^{T\!\times d}}$, respectively.

\textbf{spatiotemporal Block.} To enable UniTS to collaboratively model spatiotemporal dependencies, we introduce a spatiotemporal block with interleaved spatial and temporal attention, which captures geographic relationships in the spatial dimension and evolutionary patterns along the temporal dimension. The spatiotemporal block is structured sequentially from a spatial block to a temporal block, with each block consisting of ACor, adaptive layer normalization (AdaLN), and a spatial/temporal attention module incorporating STM, as illustrated in Fig.\ref{fig:UniTS}(a). First, ${{\bf{z}}_t}$ and ${{\bf{z}}_\text{con}}$ are reshaped into $\left\{ {{\bf{z}}_t^\text s,{\bf{z}}_\text{con}^\text s} \right\} \subset\mathbb{R}{^{T\!\times\!{n_\text h}{n_\text w}\!\times\!d}}$ as input to the spatial block. In ACor, ${\bf{z}}_\text{con}^\text s$ is injected into ${{\bf{z}}_t^\text s}$ along the spatial dimension via condition feature-wise affine transformation, with the geographic embedding ${\bf{z}}_{\text{\tiny lonlat}}$ applied. Subsequently, the spatial flow time embedding ${\bf{z}}_{\scriptscriptstyle\text{FM}}^\text s$ is mapped through a linear layer to obtain scale $\gamma_\text s$, shift $\beta_\text s$, and gating parameter $\alpha_\text s$. The parameters $\gamma_\text s$ and $\beta_\text s$ are fed into AdaLN to adjust the feature flow in the model under the guidance of flow time. The adjusted features are passed into the spatial attention module, and then the output obtained by applying gating parameter is added to the input ${\bf{z}}_t^\text s$ through a residual connection to get ${\bf{\hat z}}_t^\text s$. The workflow of the spatial block is formulated as follows,
\begin{equation}\label{eq:4}
{\bf{\hat z}}_t^{\rm{s}} = {\alpha _{\rm{s}}}\!\cdot\!{\rm{SMSA}}\left[ {{\gamma _{\rm{s}}}\!\cdot\!{\rm{LN}}\left( {{\rm{Acor}}\left( {{\bf{z}}_t^{\rm{s}},{\bf{z}}_{{\text{con}}}^{\rm{s}}} \right){\rm{ + }}{{\bf{z}}_{{\text{\tiny lonlat}}}}} \right){\rm{ + }}{\beta _{\rm{s}}}} \right]{\rm{ + }}{\bf{z}}_t^{\rm{s}}
\end{equation}
where $\rm{LN}( \cdot )$ and $\rm{SMSA}( \cdot )$ represent layer normalization and spatial multi-head self-attention module, respectively.

This ${\bf{\hat z}}_t^\text s$ containing spatial information and ${{\bf{z}}_\text{con}}$ are reshaped into $\left\{ {{\bf{z}}_t^\text t,{\bf{z}}_\text{con}^\text t} \right\} \subset\mathbb{R}{^{{n_\text h}{n_\text w}\!\times\!T\!\times\!d}}$ as input to the temporal block, where the superscript $\text t$ (also in what follows) is just a notation for a reshaped variable. Similar to the spatial block workflow described above, ${\bf{z}}_\text{con}^\text t$ is injected into ${{\bf{z}}_t^\text t}$ along the temporal dimension via affine transformation in ACor, with the date embedding ${\bf{z}}_{\text{\tiny DOY}}$ applied. 
The parameters $\gamma_\text t$, $\beta_\text t$, and $\alpha_\text t$ obtained from the temporal flow time embedding $\mathbf{z}_{\scriptscriptstyle\text{FM}}^\text t$ are input to AdaLN, and then the adjusted features are fed into the temporal attention module to obtain $\mathbf{\hat{z}}_t^\text t$. The workflow of the temporal block is formulated as follows,
\begin{equation}\label{eq:5}
{\bf{\hat z}}_t^{\text{t}} = {\alpha_{\text{t}}}\!\cdot\!{\rm{TMSA}}\left[ {{\gamma _{\text{t}}}\!\cdot\!{\rm{LN}}\left( {{\rm{Acor}}\left( {{\bf{z}}_t^{\text{t}},{\bf{z}}_{{\text{con}}}^{\text{t}}} \right){ + }{{\bf{z}}_{{\text{\tiny DOY}}}}} \right){+}{\beta _{\text{t}}}} \right]{+}{\bf{z}}_t^{\text{t}}
\end{equation}
where $\rm{TMSA}( \cdot )$ represents temporal multi-head self-attention module. After the multi-layer spatiotemporal block, we obtain the generative time series clips \({\bf{\hat x}}_t\) at flow time $t$ by using a linear decoder with AdaLN and the unpatchify operation.

\subsection{Adaptive Condition Injector (ACor)}
\label{subsec:ACor}
In generative models, effectively integrating conditional information into the model is crucial for achieving high-quality controllable generation. While the existing condition fusion strategies, particularly the metohd based on cross-attention mechanisms, have achieved success in many fields, they suffer from limitations when handling long spatiotemporal sequences, where coarse-grained fusion leads to the loss of local detail information. Inspired by adaptive normalization layers, we design ACor to embed task-specific conditions in spatial and temporal dimensions via affine transformations based on condition features. Given a feature map ${\bf{h}}_{\scriptscriptstyle\text{ACor}}$ and a condition feature ${\bf{q}}_{\scriptscriptstyle\text{ACor}}$, the condition feature ${\bf{q}}_{\scriptscriptstyle\text{ACor}}$ is passed through a convolutional layer to generate affine transformation parameters $\gamma_{\scriptscriptstyle\text{ACor}}$ and $\beta_{\scriptscriptstyle\text{ACor}}$. These parameters are used to scale and shift the feature map ${\bf{h}}_{\scriptscriptstyle\text{ACor}}$ after group normalization,

\begin{equation}\label{eq:6}
{\bf{\hat h}}_{\scriptscriptstyle\text{ACor}} = {\gamma_{\scriptscriptstyle\text{ACor}}} \cdot \rm{GN}\left( {\bf{h}}_{\scriptscriptstyle\text{ACor}} \right) + {\beta_{\scriptscriptstyle\text{ACor}}} + {\bf{h}}_{\scriptscriptstyle\text{ACor}}
\end{equation}
where $\rm{GN}( \cdot )$ represents group normalization. ACor ensures that condition feature can directly and adaptively influence the statistics of feature map ${\bf{h}}_{\scriptscriptstyle\text{ACor}}$, providing an effective integration mechanism for injecting task-specific conditions into UniTS. ACor has two variants as shown in Fig.\ref{fig:UniTS}(b):  
\begin{itemize}
	
	\item \textbf{Spatial ACor}: ${\bf{z}}_t^\text s$ and ${{\bf{z}}_\text {con}^\text s}$ are treated as the feature map ${\bf{h}}_{\scriptscriptstyle\text{ACor}}^\text s$ and condition feature ${\bf{q}}_{\scriptscriptstyle\text{ACor}}^\text s$, reshaped into the form $ T\!\times\!d\!\times\! {n_\text h}\!\times\!{n_\text w} $ and fed into ACor. A 2D convolution is applied along the spatial dimensions to generate affine transformation parameters $\gamma_{\scriptscriptstyle\text{ACor}}^\text s$ and $\beta_{\scriptscriptstyle\text{ACor}}^\text s$, which are then applied to the feature map ${\bf{h}}_{\scriptscriptstyle\text{ACor}}^\text s$. Finally, ${{\bf{\hat h}}_{\scriptscriptstyle\text{ACor}}^\text s}$ is reshaped back to the original dimensions of ${\bf{h}}_{\scriptscriptstyle\text{ACor}}^\text s$ and passed to the spatial attention module.

	\item \textbf{Temporal ACor}: ${\bf{z}}_t^\text t$ and ${{\bf{z}}_\text{con}^\text t}$ are treated as the feature map ${\bf{h}}_{\scriptscriptstyle\text{ACor}}^\text t$ and condition feature ${\bf{q}}_{\scriptscriptstyle\text{ACor}}^\text t$, reshaped into the form $ {{n_\text h}{n_\text w}\!\times\!d\!\times\!T} $ and fed into ACor. A 1D convolution is applied along the temporal axis to generate parameters $\gamma_{\scriptscriptstyle\text{ACor}}^\text t$ and $\beta_{\scriptscriptstyle\text{ACor}}^\text t$. After obtaining ${{\bf{\hat h}}_{\scriptscriptstyle\text{ACor}}^\text t}$, it is reshaped back to the original dimensions and passed to the temporal attention module.
\end{itemize}

\begin{figure*}[tp] \small
	\vspace{-4em}
	\begin{adjustwidth}{-1.4cm}{-1.4cm}
		\begin{center}
			\epsfig{width=2.4\figurewidth,file=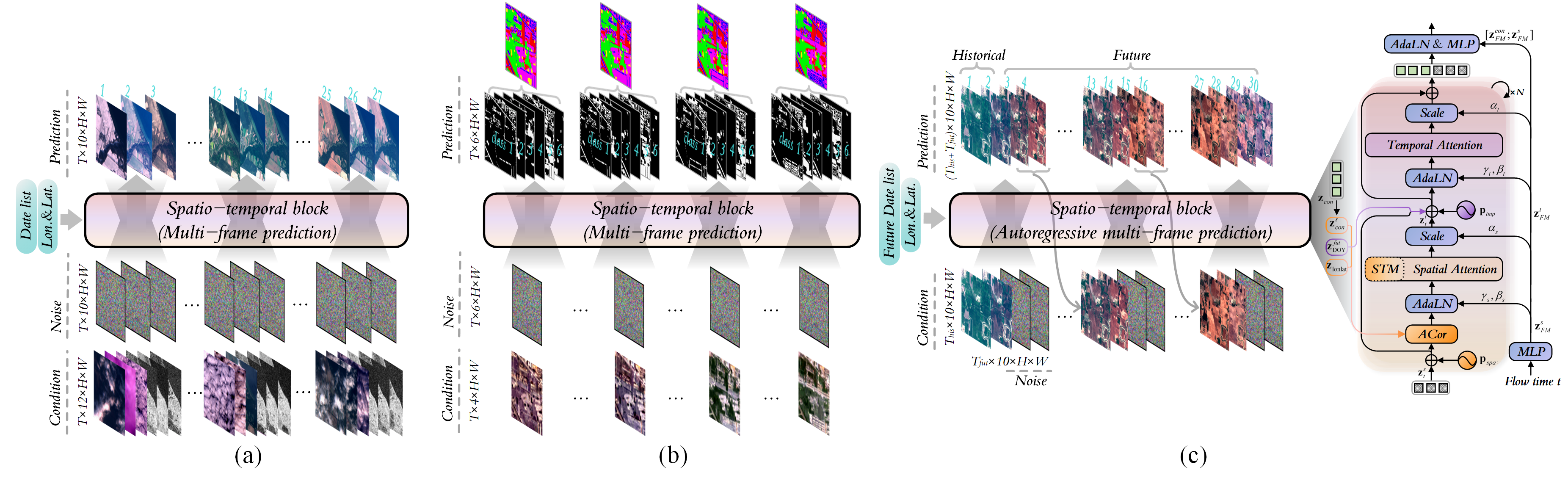}
		\end{center}
	\end{adjustwidth}
	\caption{\label{fig:inference}
		UniTS inference workflows for different time series tasks: (a) Multi-frame prediction for time series reconstruction and time series cloud removal, (b) Multi-frame prediction for time series semantic change detection, where class-specific maps are generated for each output frame, (c) Autoregressive multi-frame prediction for time series forecasting. Historical sequences and random noise are jointly fed into UniTS to predict the initial future frame. The predicted frames are then recursively used as conditions for subsequent time steps, progressively generating the full future sequence. To align with the temporal evolution characteristics of forecasting tasks, the input and spatiotemporal block are correspondingly adapted during both training and inference.}
\end{figure*}

\subsection{Spatiotemporal-aware Modulator (STM)}
\label{subsec:STM}
STM is designed to enhance the UniTS's ability to capture complex spatiotemporal dependencies. By leveraging spatiotemporal prior information embedded in auxiliary data (e.g. Sentinel-1 unaffected by cloud cover), it generates a dynamic attention bias term that directly modulates the attention weights in the spatial/temporal attention module, thereby guiding UniTS to focus on regions with higher relevance in the spatiotemporal dimensions, as shown in Fig.\ref{fig:UniTS}(c). Given auxiliary data ${\bf{q}}_{\scriptscriptstyle\text{STM}}$, we treat it as a dense, task-related prior signal (e.g., Sentinel-1 captures backscattering characteristics of different land cover types). STM encodes both the absolute positional relationship of feature map ${\bf{h}}_{\scriptscriptstyle\text{STM}}$ in flow matching and the latent relative geometric/evolutive relationship in the auxiliary data, constructing a learnable bias matrix \( \mathbf{M}_\text{spa} \in \mathbb{R}^{n_\text hn_\text w\!\times\!n_\text h n_\text w} \) (for spatial attention) or \( \mathbf{M}_\text{tmp} \in \mathbb{R}^{T\!\times\!T} \) (for temporal attention). This bias matrix is then injected into the attention score calculation as follows,
\begin{equation}\label{eq:7}
\rm{SMSA/TMSA} = {\rm{Softmax}}\left( {\frac{{{\bf{Q}}{{\bf{K}}^T}}}{\sqrt {{d}_k} } + {{\bf{M}}_\text{spa/tmp}}} \right){\bf{V}}
\end{equation}
where \( \bf{Q}, \bf{K}, \bf{V} \) represent the query, key, and value in the attention mechanism, and \( d_\text k \) denotes the feature dimension of \( \bf{K} \). This approach extends beyond content-similarity-based attention mechanisms by explicitly integrating task-relevant structural priors through STM.

STM includes two variants to handle structural dependencies in spatial dimension and dynamic evolution over temporal dimension, respectively. Auxiliary data ${{\bf{q}}_{\scriptscriptstyle\text{STM}}} \in\mathbb{R}{^{T\!\times\!{C_{\scriptscriptstyle\text{STM}}}\!\times\!{n_\text h}\!\times\!{n_\text w}}}$ uses Sentinel-1 in time series reconstruction and time series cloud removal tasks ($C_{\scriptscriptstyle\text{STM}}=2$), and uses NIR or RGB in time series semantic change detection and time series forecasting tasks ($C_{\scriptscriptstyle\text{STM}}=1{ } \text{or} { }3$).

\begin{itemize}
	
	\item \textbf{Spatial STM}:	${\bf{z}}_t^\text s$ and ${{\bf{q}}_{\scriptscriptstyle\text{STM}}}$ are regarded as feature map ${\bf{h}}_{\scriptscriptstyle\text{STM}}^\text s$ and spatial auxiliary data ${{\bf{q}}_{\scriptscriptstyle\text{STM}}^\text s}$. First, we obtain the spatial positional prior encoding by computing the Manhattan distances between all \( n_\text h \times n_\text w \) patches of ${\bf{h}}_{\scriptscriptstyle\text{STM}}^\text s$, yielding a bias matrix \( \mathbf{M}_\text {pos}^\text s \) based on absolute 2D coordinates. This models the absolute positional relationships in spatial dimension. The auxiliary data ${{\bf{q}}_{\scriptscriptstyle\text{STM}}^\text s}$ is downsampled to a spatial size of \( n_\text h \times n_\text w \). We then compute the feature differences across all spatial patches to derive the spatial auxiliary prior encoding \( \mathbf{M}_\text{aux}^\text s \), thereby capturing geometric proximity based on the auxiliary data. Using learnable weights \( w_1^\text s \) and \( w_2^\text s \), these two priors are integrated as follows,
	\begin{equation}\label{eq:8}
	{{\bf{M}}_\text{spa}} = {w_1^\text s} \cdot {\bf{M}}_\text{pos}^\text s{\rm{ }} + {w_2^\text s} \cdot {\bf{M}}_\text{aux}^\text s{\rm{ }}
	\end{equation}
	Then \( \mathbf{M}_\text{spa} \) is incorporated into Eq.(\ref{eq:7}) to modulate the spatial attention map.
	
	\item \textbf{Temporal STM}: ${\bf{z}}_t^\text t$ is regarded as feature map ${\bf{h}}_{\scriptscriptstyle\text{STM}}^\text t$, and ${{\bf{q}}_{\scriptscriptstyle\text{STM}}}$ is reshaped into the form $ {{n_\text h}{n_\text w}\!\times\!d\!\times\!T} $ as temporal auxiliary data ${{\bf{q}}_{\scriptscriptstyle\text{STM}}^\text t}$. We compute the Manhattan distance between all \( T\!\times\!T \) time steps to obtain the \( T\!\times\!T \) bias matrix \( \mathbf{M}_\text{pos}^\text t \), which models the absolute temporal distance in the time series. Additionally, we calculate the feature differences between all temporal frames of the auxiliary data along the time dimension to derive the temporal auxiliary prior encoding \( \mathbf{M}_\text{aux}^\text t \), capturing dynamic changes between temporal frames based on the characteristics of the auxiliary data. Finally, the two priors are integrated using weights \( w_1^\text t \) and \( w_2^\text t \), 
	\begin{equation}\label{eq:9}
	{{\bf{M}}_\text{tmp}} = {w_1^\text t} \cdot {\bf{M}}_\text{pos}^{\text t}{\rm{ }} + {w_2^\text t} \cdot {\bf{M}}_\text{aux}^\text t{\rm{ }}
	\end{equation}
	The resulting \( \mathbf{M}_\text{spa} \) is ultimately used to modulate the temporal attention map.
\end{itemize}

\subsection{Training and Inference for Conditional Time Series Generation}
\label{subsec:Traintest}

\textbf{Training.} During training, the input conditions, metadata, and target samples for different time series tasks are summarized in Table~\ref{tab:input}. All tasks adopt a sequence-to-sequence input-output format, where both the noise sample and the target sample have the shape ${T\!\times\!C\!\times\!H\times\!W}$. The specific settings of the training phase are as follows:
\begin{itemize}
	
	\item The outputs of time series reconstruction and time series cloud removal are identical, while their input conditions differ, the reconstruction task uses synthetic cloud-covered Sentinel-2 generated with binary cloud masks, whereas the cloud removal task uses real cloud-covered Sentinel-2 imagery. 
	\item For time series semantic change detection, the goal is to generate semantic segmentation maps for each image in the sequence, enabling long-term semantic change analysis. Here, the land cover map of each image is converted into a one-hot representation as the target sample. 
	\item In time series forecasting, the model uses historical sequences of length $T_{\text{his}}$ as conditions to predict future sequences of length $T_{\text{fut}}$, with $T_{\text{his}} = T_{\text{fut}}$ in this paper. Unlike the previous tasks, forecasting requires jointly modeling both condition and target samples during training to capture their temporal dependencies. To prevent historical patterns from interfering with future predictions, we adjust the input and spatiotemporal block architecture, as shown in Fig. \ref{fig:inference}. Specifically, the date embedding ${\bf{z}}_{\text{\tiny DOY}}^{\scriptscriptstyle\text{fut}} \in\mathbb{R}{^{{T_{\scriptscriptstyle\text{fut}}} \times d}}$ is generated using only future dates from the metadata. A historical condition timestep embedding ${\bf{z}}_{\scriptscriptstyle\text{FM}}^\text{con} \in\mathbb{R}{^{{T_{\scriptscriptstyle\text{his}}} \times d}}$ is added to the flow time embedding. In the temporal block, the ACor and STM components are removed. During training, ${\bf{z}}_{\text{\tiny DOY}}^{\scriptscriptstyle\text{fut}}$ is applied to the feature flow before the temporal block. After passing through spatiotemporal blocks, the condition tokens $\mathbf{z}_{\text{con}}$ and $\hat{\mathbf{z}}_t^{\text{t}}$ are concatenated, along with ${\bf{z}}_{\scriptscriptstyle\text{FM}}^\text{con}$ and ${\bf{z}}_{\scriptscriptstyle\text{FM}}^\text s$, and fed into a linear decoder with AdaLN.

\end{itemize}

\textbf{Inference.} As illustrated in Fig.\ref{fig:inference}, UniTS inference workflows for different time series tasks include multi-frame prediction for time series reconstruction, time series cloud removal, and time series semantic change detection. For time series forecasting, an autoregressive multi-frame prediction approach is employed. UniTS takes both historical sequences and random noise as joint input to generate initial future sequences predictions. These predicted sequences are then recursively used as conditions for subsequent time steps, progressively generating the complete future sequences.

\section{Experimental results and analysis}
\label{sec:results}


\subsection{Implementation Details}

We follow the DiT architecture to implement UniTS. Unlike latent transformers that rely on pre-trained VAEs to compress video or time series inputs, UniTS directly tokenizes the pixel space. This is because most inputs to UniTS are multispectral images with significantly more channels than three, for which existing pre-trained VAE models are unsuitable. All experiments are trained using the AdamW optimizer with a constant learning rate of $1e-4$. The input image size is fixed at $128 \times 128$, and the Dopri5 solver is employed for sampling with 10 sample steps. The model, training hyperparameters, and inference settings are provided in Appendix A. The qualitative comparison of all experiments are provided in Appendix C-G.

\subsection{Time Series Reconstruction}
\textbf{Dataset and Evaluation Settings.} We conduct benchmark tests on time series reconstruction using the TS-S12 dataset, which comprises 10,481 training samples and 4,492 test samples, with sequence lengths ranging from 8$\sim$97. The model is trained in a sequence-to-sequence manner, where both the conditional and target sample sequences are set to a fixed length of $T=8$. During inference, all test samples are processed sequentially with a sliding window of length 8, and each sample is reasoned until the complete sequence length of that sample is generated. The comparison methods in the experiment include time series Interpolator (Last, Closest, Linear interpolation), non-generative model (U-TILISE \cite{10319805}, UnCRtainTS \cite{10209044} and VRT \cite{liang2024vrt}), and generative model (EMRDM \cite{liu2025effective}, SiT \cite{ma2024sit}, VDPS \cite{kwon2025video}, SeedVR \cite{wang2025seedvr}, and RESTORE-DiT \cite{shu2025restore}). To comprehensively evaluate performance, we employ multiple metrics, including Peak Signal-to-Noise Ratio (PSNR), Structural Similarity Index (SSIM), Root Mean Square Error (RMSE), Mean Absolute Error (MAE), and Spectral Angle Mapper (SAM).

\begin{table*}[]
	\vspace{-4em}
	\centering
	\caption{ Quantitative Comparison of Time Series Reconstruction on TS-S12 Dataset. For full-band evaluation, the best and second-best values are bolded and underlined respectively. S1-Sentinel-1, S2-Sentinel-2.}
	\label{tab:cldmsk}
	\begin{tabular}{c|c|c|ccccc|c}
		\toprule
		Methods              & Conditions        & Time series   & PSNR↑ & SSIM↑        & RMSE↓  & MAE↓   & SAM↓ & Params \\ \midrule
		\multicolumn{9}{c}{\cellcolor[HTML]{E7E7E7}\textit{\textbf{Time series Interpolator}}}                                                                       \\ \midrule
		Last                 & Synthetic cloud-covered S2        & $\checkmark$ & 17.26 & 0.8685       & 0.1549 & 0.0624 & 3.94 &    -       \\
		Closest              & Synthetic cloud-covered S2        & $\checkmark$ & 18.21 & 0.8914       & 0.1279 & 0.0545 & 3.76 &    -       \\
		Linear interpolation & Synthetic cloud-covered S2        & $\checkmark$ & 20.87 & 0.9039       & 0.1195 & 0.0521 & 3.57 &     -      \\ 
		\midrule
		\multicolumn{9}{c}{\cellcolor[HTML]{E7E7E7}\textit{\textbf{Non-generative model}}}                                                                       \\ \midrule
		U-TILISE \cite{10319805}            & Synthetic cloud-covered S2 \& S1 & $\checkmark$ & 28.69 & 0.9157       & 0.0393 & 0.0251 & 4.13 & 1.36M     \\
		UnCRtainTS \cite{10209044}          & Synthetic cloud-covered S2 \& S1 & $\checkmark$ & 21.83 & 0.8536       & 0.0859 & 0.0484 & 6.71 & 0.52M     \\
		VRT \cite{liang2024vrt}                 & Synthetic cloud-covered S2        & $\checkmark$ & 23.63 & 0.8722       & 0.0686 & 0.0456 & 5.42 & 10.26M    \\ 
		\midrule
		\multicolumn{9}{c}{\cellcolor[HTML]{E7E7E7}\textit{\textbf{Generative model}}}                                                                       \\ \midrule
		EMRDM \cite{liu2025effective}               & Synthetic cloud-covered S2 \& S1 & ×            & 25.96 & 0.8861       & 0.0518 & 0.0403 & 4.69 & 39.22M     \\
		SiT \cite{ma2024sit}                 & Synthetic cloud-covered S2 \& S1 & ×            & 26.01 & 0.8629       & 0.0522 & 0.0386 & 5.22 & 45.08M    \\
		VDPS \cite{kwon2025video}                & Synthetic cloud-covered S2        & $\checkmark$ & 21.66 & 0.8542       & 0.0898 & 0.0501 & 6.97 & 40.57M    \\
		SeedVR \cite{wang2025seedvr}              & Synthetic cloud-covered S2 \& S1 & $\checkmark$ & 29.06 & {\underline {0.9187}} & 0.0381 & 0.0204 & 3.52 & 59.05M    \\
		RESTORE-DiT \cite{shu2025restore}         & Synthetic cloud-covered S2 \& S1 & $\checkmark$ & 28.44 & 0.9121       & 0.0402 & 0.0232 & 3.97 & 10.12M    \\
		UniTS & Synthetic cloud-covered S2        & $\checkmark$ & {\underline {29.43}}    & 0.9143          & {\underline {0.0376}}    & {\underline {0.0191}}    & {\underline {3.18}}    & 54.72M \\
		\cellcolor[HTML]{E6E6FA}{UniTS} & \cellcolor[HTML]{E6E6FA}{Synthetic cloud-covered S2 \& S1} & \cellcolor[HTML]{E6E6FA}{$\checkmark$} & \cellcolor[HTML]{E6E6FA}{\textbf{30.15}} & \cellcolor[HTML]{E6E6FA}{\textbf{0.9261}} & \cellcolor[HTML]{E6E6FA}{\textbf{0.0358}} & \cellcolor[HTML]{E6E6FA}{\textbf{0.0159}} & \cellcolor[HTML]{E6E6FA}{\textbf{3.01}} & \cellcolor[HTML]{E6E6FA}{54.75M} \\ 
		\bottomrule
	\end{tabular}
\end{table*}

\begin{figure*} \small
	\begin{center}
		\includegraphics[width=1\textwidth]{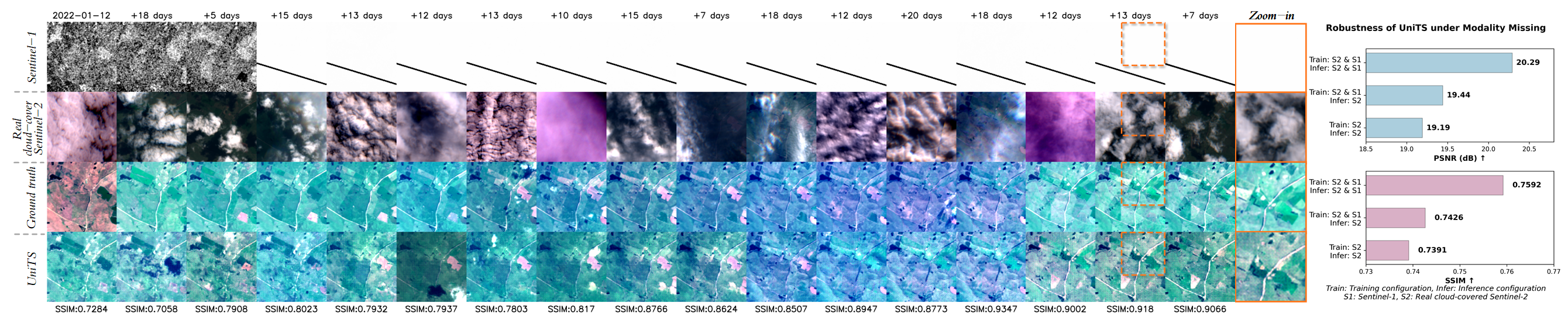}
	\end{center}
	\caption{\label{fig:rmcld-miss} Qualitative comparison of time series cloud removal under modality missing on TS-S12CR Dataset, presenting the RGB band of Sentinel-2 here. The SSIM value of each frames in the time series is marked. (Lavalleja, Uruguay, $(33^\circ47'23.3''\mathrm{S},\ 55^\circ09'25.6''\mathrm{W})$)}
		\vspace{-2em}
\end{figure*}

\begin{figure}[tp]
	\begin{center}
		\centering
		\begin{tabular}{cc}
			\epsfig{width=\figurewidth,file=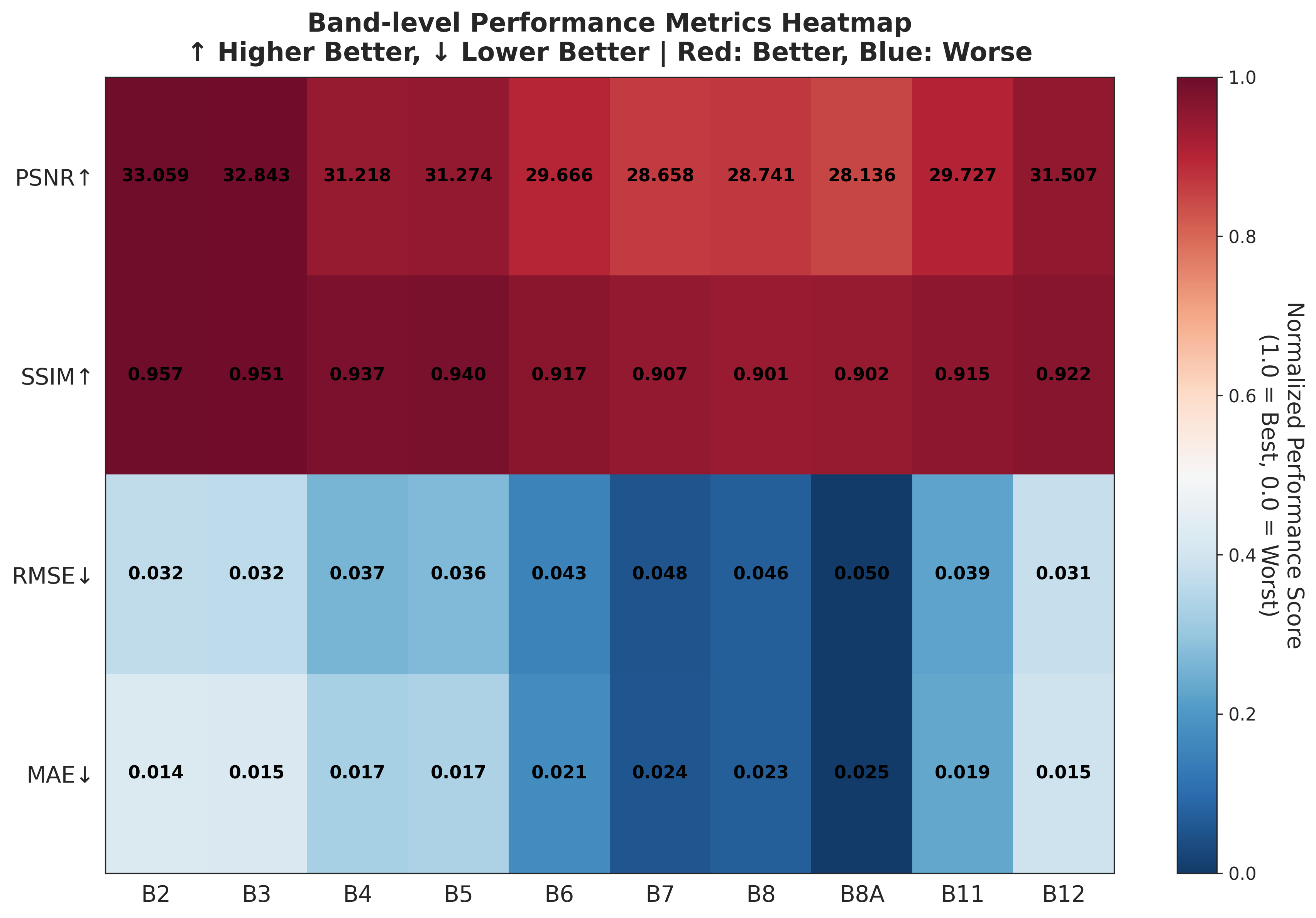} \\(a) Band-level heatmap on TS-S12 Dataset \\
			\epsfig{width=\figurewidth,file=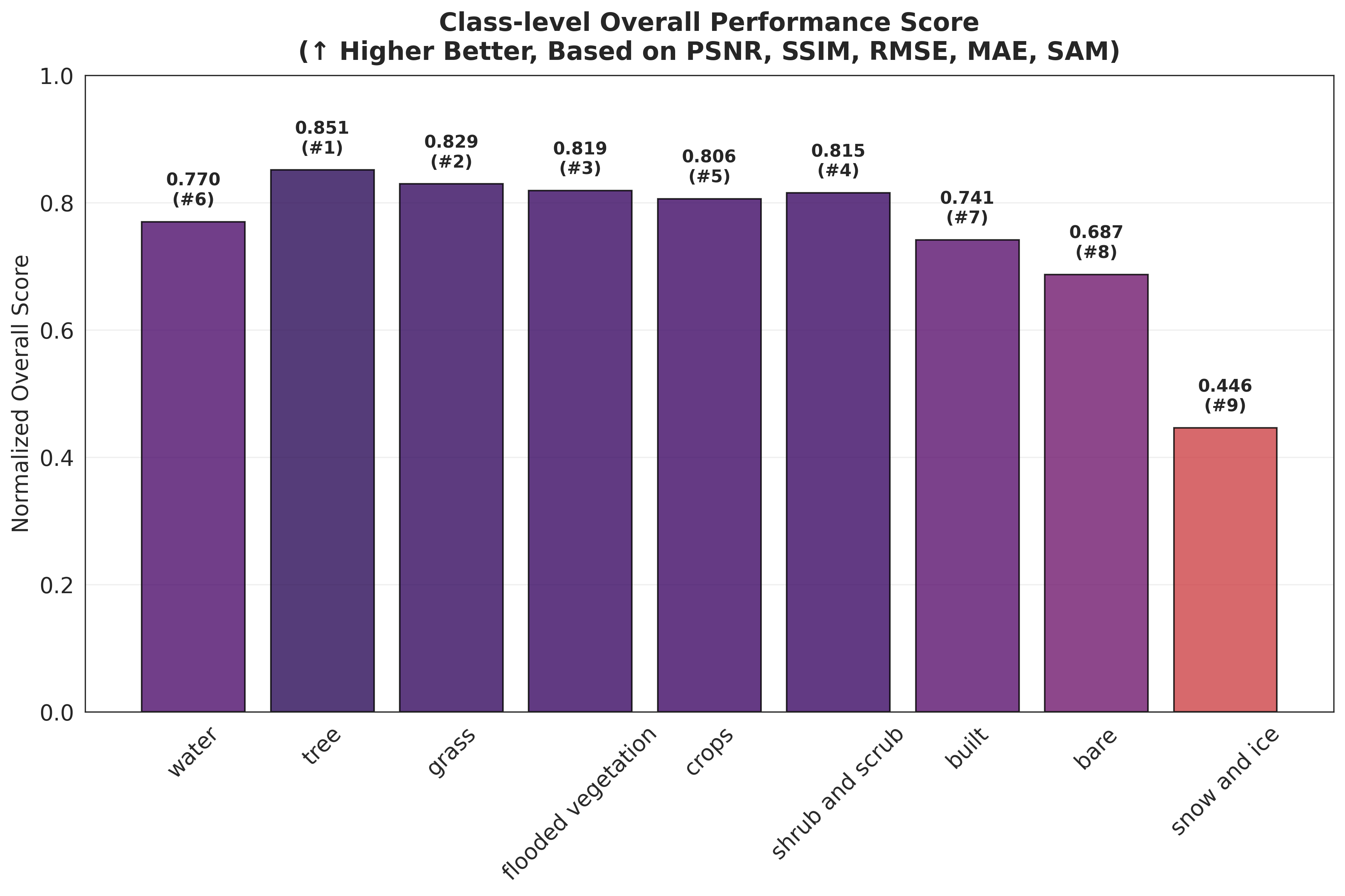}\\
			\\ (b) Class-level score on TS-S12 Dataset\\ 
		\end{tabular}
		\vspace*{0in}
		\caption{\label{fig:cldmsk_anal}
			Band-level and class-level reconstruction TS-S12 Dataset.}
	\end{center}
	\vspace{-2em}
\end{figure}

\textbf{Quantitative Comparison.} Table~\ref{tab:cldmsk} summarizes the input conditions, temporal modeling capacity, parameters, and performance of all methods on TS-S12 (all reconstructing 10 spectral bands of Sentinel-2). While conventional interpolators and video based methods (VRT, VDPS) use single-modal inputs, most other approaches support multimodal data; EMRDM and SiT, however, lack temporal modeling. UniTS significantly outperforms all existing satellite time-series reconstruction and video-restoration methods. Even with Sentinel-2 only, it leads on most metrics. Under the same multimodal setting as the strongest baseline (SeedVR), UniTS further improves PSNR by 1.09 dB and reduces SAM by 0.51, demonstrating superior time-series reconstruction and multimodal fusion.

\textbf{Band-level and Class-level Reconstruction.} We further analyze UniTS's reconstruction performance under multi-modal conditional inputs across Sentinel-2 spectral bands and land cover classes. Fig.~\ref{fig:cldmsk_anal}(a) shows band-level metrics via heatmap, while Fig.~\ref{fig:cldmsk_anal}(b) presents the class-level score averaged from five metrics. As shown in Fig.~\ref{fig:cldmsk_anal}(a), UniTS exhibits increasing RMSE from visible to near-infrared bands (B2–B8A), peaking at Red Edge 4 (B8A), then decreasing in short-wave infrared bands (B11–B12), a trend consistent with RESTORE-DiT \cite{shu2025restore}.  Among the nine land cover classes in the TS-S12 dataset, trees achieve the highest reconstruction performance (Fig.~\ref{fig:cldmsk_anal}(b)). The four vegetation types, grass, flooded vegetation, crops, and shrub/scrub, show similar scores. In contrast, snow and ice exhibit significantly lower performance due to their high spectral variability and strong reflectance, which complicates their distinction from clouds and accurate spectral modeling.

\textbf{Reconstruction under different missing ratio.} Table~\ref{tab:miss-ratio} provides the time series reconstruction performance of UniTS under different missing rates. We simulate different degrees of data missing by randomly selecting a specific proportion of frames in the input sequence and replacing them with all one cloud masks. The missing rate is defined as the proportion of the number of missing frames to the total length of the time series, Four missing conditions of 30\%, 50\%, 70\%, and 90\% are set. The experimental results show that as the missing rate increases, the difficulty of reconstruction rises sharply, and the model performance presents a downward trend.

\begin{table}[]
	\centering
	\caption{Reconstruction under different missing ratio.}
	\label{tab:miss-ratio}
	\begin{tabular}{c|ccccc}
		\toprule
		Missing rate & PSNR↑ & SSIM↑ & RMSE↓ & MAE↓ & SAM↓ \\ \midrule
		30\% & \textbf{30.75} & \textbf{0.9561} & \textbf{0.0351} & \textbf{0.0141} & \textbf{2.37} \\
		50\% & 27.97 & 0.9325 & 0.0469 & 0.0213 & 3.21 \\
		70\% & 24.43 & 0.8790 & 0.0691 & 0.0363 & 5.33 \\
		90\% & 19.08 & 0.7349 & 0.1224 & 0.0776 & 11.89 \\ \bottomrule
	\end{tabular}
\end{table}

\subsection{Time Series Cloud Removal}
\textbf{Dataset and Evaluation Settings.} We conduct benchmark tests on time series cloud removal using the TS-S12CR dataset, which comprises 8,488 training samples and 3,638 test samples, with sequence lengths ranging from 8$\sim$44. The model is trained in a sequence-to-sequence manner, where both the conditional and target sample sequences are set to a fixed length of $T=8$. The comparison methods and evaluation metrics in the experiment are consistent with those in the time series reconstruction task.

\begin{table*}[]
	\vspace{-4em}
	\centering
	\caption{ Quantitative Comparison of Time Series Cloud Removal on TS-S12CR Dataset. For full-band evaluation, the best and second-best values are bolded and underlined respectively. S1-Sentinel-1, S2-Sentinel-2.}
	\label{tab:rmcld}
	\begin{tabular}{c|c|c|ccccc|c}
		\toprule
		Methods              & Conditions        & Time series   & PSNR↑ & SSIM↑  & RMSE↓  & MAE↓   & SAM↓  & Params \\ \midrule
		\multicolumn{9}{c}{\cellcolor[HTML]{E7E7E7}\textit{\textbf{Time series Interpolator}}}                                                                       \\ \midrule
		Last                 & Real cloud-covered S2        & $\checkmark$ & 13.05 & 0.5815 & 0.2302 & 0.1789 & 16.62 &     -      \\
		Closest              & Real cloud-covered S2        & $\checkmark$ & 13.24 & 0.5924 & 0.2235 & 0.1711 & 16.06 &     -      \\
		Linear interpolation & Real cloud-covered S2        & $\checkmark$ & 12.62 & 0.5803 & 0.2415 & 0.1842 & 16.11 &      -     \\ \midrule
		\multicolumn{9}{c}{\cellcolor[HTML]{E7E7E7}\textit{\textbf{Non-generative model}}}                                                                       \\ \midrule
		U-TILISE \cite{10319805}              & Real cloud-covered S2 \& S1 & $\checkmark$ & 18.41 & 0.7267 & 0.1298 & 0.0825 & 9.81  & 1.36M     \\
		UnCRtainTS \cite{10209044}          & Real cloud-covered S2 \& S1 & $\checkmark$ & 17.86 & 0.6769 & 0.1451 & 0.1105 & 12.96 & 0.52M     \\
		VRT \cite{liang2024vrt}                 & Real cloud-covered S2        & $\checkmark$ & 16.00 & 0.6075 & 0.1744 & 0.1609 & 14.39 & 10.26M    \\ \midrule
		\multicolumn{9}{c}{\cellcolor[HTML]{E7E7E7}\textit{\textbf{Generative model}}}                                                                       \\ \midrule
		EMRDM \cite{liu2025effective}               & Real cloud-covered S2 \& S1 & ×            & 15.13 & 0.6358 & 0.1903 & 0.1652 & 11.61 & 39.22M     \\
		SiT \cite{ma2024sit}                & Real cloud-covered S2 \& S1 & ×            & 16.58 & 0.6681 & 0.1616 & 0.1241 & 10.31 & 45.08M    \\
		VDPS \cite{kwon2025video}                & Real cloud-covered S2        & $\checkmark$ & 17.65 & 0.6978 & 0.1512 & 0.1322 & 8.76  & 40.57M    \\
		SeedVR \cite{wang2025seedvr}              & Real cloud-covered S2 \& S1 & $\checkmark$ & 15.33 & 0.6423 & 0.1872 & 0.1576 & 12.07 & 59.05M    \\
		RESTORE-DiT \cite{shu2025restore}          & Real cloud-covered S2 \& S1 & $\checkmark$ & 17.01 & 0.6451 & 0.1569 & 0.1363 & 9.01  & 10.12M    \\
		UniTS & Real cloud-covered S2        & $\checkmark$ & {\underline {19.19}}    & {\underline {0.7391}}    & {\underline {0.1223}}    & {\underline {0.0890}}    & {\underline {7.83}}    & 54.72M \\
		\cellcolor[HTML]{E6E6FA}{UniTS} & \cellcolor[HTML]{E6E6FA}{Real cloud-covered S2 \& S1} & \cellcolor[HTML]{E6E6FA}{$\checkmark$} & \cellcolor[HTML]{E6E6FA}{\textbf{20.29}} & \cellcolor[HTML]{E6E6FA}{\textbf{0.7592}} & \cellcolor[HTML]{E6E6FA}{\textbf{0.1103}} & \cellcolor[HTML]{E6E6FA}{\textbf{0.0828}} & \cellcolor[HTML]{E6E6FA}{\textbf{7.42}} & \cellcolor[HTML]{E6E6FA}{54.75M} \\ \bottomrule
	\end{tabular}
\vspace{-2em}
\end{table*}

\textbf{Quantitative Comparison.} Table~\ref{tab:rmcld} shows time series cloud-removal results on TS-S12CR. UniTS significantly outperforms all existing methods, improving PSNR over 1.88 dB and SSIM over 3.25\% regardless of whether Sentinel-1 is used. This fully demonstrates the effectiveness of UniTS in handling real cloud noise. Notably, compared with the time series reconstruction results in Table~\ref{tab:cldmsk}, UniTS's PSNR drops by ~9.86 dB. \emph{This large gap highlights that the complexity of real cloud coverage scenarios far exceeds that of the simplified scenarios constructed through simulated cloud masks in current mainstream time series reconstruction or gap-filling studies.} By building TS-S12CR to closely reflect real conditions, we establish a practical benchmark for future time series cloud-removal research.

\textbf{Band-level and Class-level Cloud Removal.} We further analyze the cloud removal performance of UniTS on severely cloud-contaminated Sentinel-2 imagery across different spectral bands and land cover classes. the band-level metric heatmap and the class-level overall performance score can be found in Appendix D. The experimental conclusions at the band-level and class-level are consistent with the observation results of the time series reconstruction task.

\textbf{Robustness to Modality Absence During Inference.} Multimodal cloud-removal methods often assume all modalities are available during inference. However, sensor failures or acquisition gaps can leave certain modalities entirely missing. While approaches like U-TILISE suffer significant performance degradation when Sentinel-1 is absent or noisy, UniTS maintains stable cloud-removal quality even when Sentinel-1 is entirely absent. We evaluate robustness under different training and inference modality configurations in Fig.\ref{fig:rmcld-miss}. When trained on full modalities but tested with Sentinel-1 replaced by zero masks, UniTS shows only a moderate PSNR drop (-0.85 dB) and still outperforming the Sentinel-2-only baseline.

\subsection{Time Series Semantic Change Detection}

\begin{table*}[]
	\centering
	\caption{Quantitative Comparison of Time Series Semantic Change Detection on DynamicEarthNet Dataset. The best and second-best values are bolded and underlined respectively.}
	\label{tab:Dync}
	\setlength\tabcolsep{4pt}
	\begin{tabular}{c|c|c|cccccc|cccc|c}
		\toprule
		&                              &                            & \multicolumn{6}{c|}{Per-class IoU↑}                     & \multicolumn{4}{c|}{Overall Metrics}             &                             \\
		\multirow{-2}{*}{Methods} & \multirow{-2}{*}{Conditions} & \multirow{-2}{*}{Strategy} & \cellcolor[HTML]{696969}{imp. surf.} & \cellcolor[HTML]{9ACD32}{agr.}  & \cellcolor[HTML]{228B22}{forest} & \cellcolor[HTML]{20B2AA}{wetlands} & \cellcolor[HTML]{D2691E}{soil}  & \cellcolor[HTML]{1E90FF}{water} & mIoU↑          & BC↑   & SC↑            &  \multicolumn{1}{c|}{SCS↑}  & \multirow{-2}{*}{Params} \\ \midrule
		\multicolumn{14}{c}{\cellcolor[HTML]{E7E7E7}\textit{\textbf{Specific model}}}                                                                                                                                                                                                                                    \\ \midrule
		TSViT \cite{tarasiou2023vits}                    & Planet (RGBN)                & Mono                       & 22.28      & 12.40  & 53.23  & 0.0        & 36.91 & 70.14 & 32.49          & 3.03  & 16.29          & 9.66  & 1.72M                                                                  \\
		U-TAE \cite{garnot2021panoptic}                     & Planet (RGBN)                & Mono                       & 13.96      & 13.96 & 59.61  & 0.36     & 50.06 & 89.52 & \underline{40.94}          & 9.68  & \underline{26.63}          & 18.16 & 1.08M                                                                  \\
		A2Net \cite{li2023lightweight}                    & Planet (RGBN)                & Bi                         & \underline{26.57}      & 18.89 & 59.62  & 0.98    & 30.09 & 87.66 & 37.31          & 9.33  & 19.41          & 14.37 & 1.83M                                                                  \\
		SCanNet \cite{ding2024joint}                  & Planet (RGBN)                & Bi                         & 11.71      & \underline{20.24} & 60.75  & 0.0        & \underline{53.41} & 88.28 & 39.06          & 10.08 & 21.94          & 16.01 & 27.9M                                                                  \\
		TSSCD \cite{he2024time}                    & Planet (RGBN)                & Multi                      & 21.81      & 2.86  & 40.86  & \underline{2.81}     & 41.75 & 78.65 & 31.46          & 6.64  & 18.69          & 12.66 & 6.11M                                                                  \\ \midrule
		\multicolumn{14}{c}{\cellcolor[HTML]{E7E7E7}\textit{\textbf{Foundation model}}}                                                                                                                                                                                                                                    \\ \midrule
		Scale-MAE \cite{reed2023scale}                & Planet (RGBN)                & Mono                       & 14.01      & \textbf{23.87} & 57.11  & 0.0        & 45.92 & \underline{94.53} & 39.23          & 5.27  & 21.62          & 13.45 & 303.4M                                                                \\
		SatMAE++ \cite{noman2024rethinking}                 & Planet (RGBN)                & Mono                       & 15.42      & 16.51 & \textbf{62.21}  & 0.0        & 42.42 & \textbf{95.23} & 38.63          & 6.39  & 23.02          & 14.71 & 305.5M                                                                \\
		AnySat \cite{astruc2025anysat}                   & Planet (RGBN)                & Mono                       & 0.0          & 0.0     & 34.49  & 0.0        & 14.70  & 67.51 & 19.45          & 2.82  & 7.46           & 5.14  & 127.77M                                                                \\
		SkySense \cite{guo2024skysense}                 & Planet (RGBN)                & Mono                       & 14.12      & 15.21 & 60.91  & 0.49     & 41.12 & 92.93 & 36.79         & \textbf{13.60} & 24.66          & \textbf{19.54} & 661.99M                                                                \\
		\cellcolor[HTML]{E6E6FA}{UniTS}                     & \cellcolor[HTML]{E6E6FA}{Planet (RGBN)}                & \cellcolor[HTML]{E6E6FA}{Multi}                      & \cellcolor[HTML]{E6E6FA}{\textbf{27.89}}      & \cellcolor[HTML]{E6E6FA}{17.13} & \cellcolor[HTML]{E6E6FA}{\underline{60.97}}  & \cellcolor[HTML]{E6E6FA}{\textbf{4.31}}     & \cellcolor[HTML]{E6E6FA}{\textbf{54.29}} & \cellcolor[HTML]{E6E6FA}{90.61} & \cellcolor[HTML]{E6E6FA}{\textbf{42.52}} & \cellcolor[HTML]{E6E6FA}{\underline{12.45}} & \cellcolor[HTML]{E6E6FA}{\textbf{27.41}} & \cellcolor[HTML]{E6E6FA}{\underline{18.43}} & \cellcolor[HTML]{E6E6FA}{54.21M}  \\           \bottomrule                                                   
	\end{tabular}
\end{table*}

\begin{table}[]
	\vspace{-1em}
	\centering
	\caption{Quantitative Comparison of Time Series Semantic Change Detection on MUDS Dataset. The best and second-best values are bolded and underlined respectively.}
	\label{tab:MUDS}
	\setlength\tabcolsep{1pt}
	\scriptsize
	\begin{tabular}{c|c|c|cc|cccc|c}
		\toprule
		\multirow{2}{*}{Methods} & \multirow{2}{*}{Cond.} & \multicolumn{1}{c|}{\multirow{2}{*}{Str.}} & \multicolumn{2}{c|}{Per-class IoU↑}          & \multicolumn{4}{c|}{Overall Metrics}                                        & \multirow{2}{*}{Params} \\
		&                             & \multicolumn{1}{c|}{}                          & not build.     & \multicolumn{1}{c|}{build.} & mIoU↑          & BC↑           & SC↑            & \multicolumn{1}{c|}{SCS↑} &                            \\ \midrule
		\multicolumn{10}{c}{\cellcolor[HTML]{E7E7E7}\textit{\textbf{Specific model}}}                                                                                                                                                                                                                                    \\ \midrule
		TSViT \cite{tarasiou2023vits}                   & Planet (RGB)                & Mono                                           & 87.15          & 16.41                       & 51.78          & 0.13          & 9.54           & 4.84                      & 1.72M                       \\
		UTAE \cite{garnot2021panoptic}                    & Planet (RGB)                & Mono                                           & 88.95          & \underline{32.52}                       & \underline{60.74}          & 0.36          & 24.86          & 12.61                     & 1.08M                       \\
		A2Net \cite{li2023lightweight}                   & Planet (RGB)                & Bi                                             & 81.37          & 28.07                       & 54.72          & 0.27          & \underline{38.22}          & \underline{19.24}                     & 1.83M                       \\
		SCanNet \cite{ding2024joint}                 & Planet (RGB)                & Bi                                             & 87.51          & 30.94                       & 59.22          & 0.35          & 28.61          & 14.47                     & 27.9M                       \\
		TSSCD \cite{he2024time}                   & Planet (RGB)                & Multi                                          & 85.87          & 11.42                       & 48.64          & 0.18          & 13.81          & 6.99                      & 6.11M                       \\ \midrule
		\multicolumn{10}{c}{\cellcolor[HTML]{E7E7E7}\textit{\textbf{Foundation model}}}                                                                                                                                                                                                                                    \\ \midrule
		Scale-MAE \cite{noman2024rethinking}               & Planet (RGB)                & Mono                                           & \textbf{94.91} & 16.55                       & 55.73          & 1.44          & 30.21          & 16.32                     & 303.4M                     \\
		SatMAE++ \cite{noman2024rethinking}                & Planet (RGB)                & Mono                                           & \underline{94.81}          & 11.72                       & 53.26          & \underline{1.95}          & 27.96          & 15.94                     & 305.5M                     \\
		AnySat \cite{astruc2025anysat}                  & Planet (RGB)                & Multi                                           & 84.33          & 24.73                       & 54.53          & 0             & 0              & 0                         & 127.77M                     \\
		SkySense \cite{guo2024skysense}                 & Planet (RGB)                & Mono                                           & 87.03          & 27.43                       & 57.23          & 0.29          & 33.63          & 16.96                     & 661.99M                     \\
		\cellcolor[HTML]{E6E6FA}{UniTS}                    & \cellcolor[HTML]{E6E6FA}{Planet (RGB)}                & \cellcolor[HTML]{E6E6FA}{Multi}                                          & \cellcolor[HTML]{E6E6FA}{91.81}          & \cellcolor[HTML]{E6E6FA}{\textbf{33.18}}              & \cellcolor[HTML]{E6E6FA}{\textbf{61.96}} & \cellcolor[HTML]{E6E6FA}{\textbf{8.54}} & \cellcolor[HTML]{E6E6FA}{\textbf{49.81}} & \cellcolor[HTML]{E6E6FA}{\textbf{29.17}}            & \cellcolor[HTML]{E6E6FA}{54.21M}     \\ \bottomrule                
	\end{tabular}
\end{table}



\textbf{Dataset and Evaluation Settings.} We evaluate our method on DynamicEarthNet \cite{toker2022dynamicearthnet} and MUDS \cite{van2021multi}. DynamicEarthNet contains monthly Planet satellite images from 2018–2019, with 54,750 RGBN images of size 1024×1024 and 3-meter resolution, annotated for land use and land cover. MUDS includes monthly Planet RGB images from 101 global regions in the same period, annotated with building footprints. A sequence-to-sequence training and inference strategy is adopted, where both the conditional and target sample sequences are set to a a fixed length of $T=6$. The experiments compare against several state-of-the-art semantic change detection models (TSViT \cite{tarasiou2023vits}, U-TAE \cite{garnot2021panoptic}, A2Net \cite{li2023lightweight}, SCanNet \cite{ding2024joint} and TSSCD \cite{he2024time}) and multiple foundational remote sensing models (Scale-MAE \cite{reed2023scale}, SatMAE++ \cite{noman2024rethinking}, AnySat \cite{astruc2025anysat} and Skysense \cite{guo2024skysense}). We employ multiple metrics to evaluate the performance, including the mean intersection-over-union (mIoU), binary change score (BC), semantic change score (SC) and semantic change segmentation score (SCS).

\textbf{Quantitative Comparison.} Table~\ref{tab:Dync} and Table~\ref{tab:MUDS} present the per-class IoU and overall metrics of all methods on the two datasets. The strategy column indicates the input-output setup: \emph{Mono} uses a single frame to produce one segmentation map; \emph{Bi} takes two temporal images and outputs a binary change mask plus two semantic maps; \emph{Multi} takes the full time series and generates segmentation for each frame. UniTS achieves the best interpretation performance on both datasets, with mIoU scores of 42.52\% (DynamicEarthNet) and 61.96\% (MUDS). These results highlight the effectiveness of UniTS’s spatiotemporal block in capturing spatiotemporal dependencies and performing precise temporal semantic segmentation. They also demonstrate that representations learned through flow matching generative objectives retain strong generalization and spatiotemporal reasoning ability without requiring large‑scale pretraining.


\subsection{Time Series Forecasting}

\textbf{Dataset and Evaluation Settings.} We evaluate the performance of time series forecasting on TS-S12 and GreenEarthNet \cite{10656834}. GreenEarthNet contains high-resolution vegetation sequences from Europe, each with 30 frames (10 historical, 20 future) at 5-day intervals, 4 Sentinel-2 bands (B2, B3, B4, B8), daily meteorological observations, and an elevation map. Its training/test sets include 23,816 and 4,205 samples, respectively. Note that the multimodal conditions in GreenEarthNet have inconsistent temporal dimensions; detailed fusion strategies are provided in Appendix B. Compared methods include: video prediction models (STAU \cite{chang2025stau}, Latte \cite{ma2025latte}, SyncVP \cite{pallotta2025syncvp}, LARP \cite{wang2024larp}, DFoT \cite{song2025history}, FAR \cite{gu2025long}); and for GreenEarthNet, additional non-generative baselines (PredRNN \cite{wang2022predrnn}, SimVP \cite{tan2025simvpv2}, Contextformer \cite{10656834}). Nongenerative methods follow sequence-to-sequence training/inference; generative methods use sequence-to-sequence training but autoregressive multi-frame prediction at inference. Historical sequence length ${T_\text{his}}$ is 4 (TS-S12) and 10 (GreenEarthNet). Metrics follow the reconstruction task.

\begin{table}[]
	\vspace{-1em}
	\centering
	\caption{ Quantitative Comparison of Time Series Forecasting on TS-S12 Dataset. For full-band evaluation, the best and second-best values are bolded and underlined respectively. S2-Sentinel-2.}
	\label{tab:pred}
		\setlength\tabcolsep{2pt}
	\scriptsize
	\begin{tabular}{c|c|c|ccccc|c}
		\toprule
		Methods & Cond.            & Time. & PSNR↑          & SSIM↑           & RMSE↓           & MAE↓            & SAM↓          & Params  \\ \midrule
		\multicolumn{9}{c}{\cellcolor[HTML]{E7E7E7}\textit{\textbf{Non-generative model}}}                                                                                                                                                                                                                                    \\ \midrule
		STAU \cite{chang2025stau}   & Historical S2 & $\checkmark$           & 20.39          & 0.7373          & 0.1198          & 0.0898          & 12.32         & 7.63M  \\
		\midrule
		\multicolumn{9}{c}{\cellcolor[HTML]{E7E7E7}\textit{\textbf{Generative model}}}                                                                                                                                                                                                                                    \\ \midrule
		Latte \cite{ma2025latte}  & Historical S2 & $\checkmark$           & 18.33          & 0.6582          & 0.1564          & 0.1313          & 12.24         & 32.32M \\
		SyncVP \cite{pallotta2025syncvp} & Historical S2 & $\checkmark$           & \underline{21.05}          & \underline{0.7701}          & \underline{0.0978}          & \underline{0.0711}          & \underline{9.93}          & 58.56M \\
		LARP \cite{wang2024larp}   & Historical S2 & $\checkmark$           & 17.21          & 0.6299          & 0.1595          & 0.1294          & 14.45         & 27.97M \\
		DFoT \cite{song2025history}   & Historical S2 & $\checkmark$           & 16.09          & 0.6336          & 0.1844          & 0.1543          & 15.55         & 32.88M \\
		FAR \cite{gu2025long}    & Historical S2 & $\checkmark$           & 16.23          & 0.7212          & 0.1581          & 0.1183          & 10.71         & 32.59M \\
		\cellcolor[HTML]{E6E6FA}{UniTS}   & 	\cellcolor[HTML]{E6E6FA}{Historical S2} & 	\cellcolor[HTML]{E6E6FA}{$\checkmark$}           & 	\cellcolor[HTML]{E6E6FA}{\textbf{22.57}} & 	\cellcolor[HTML]{E6E6FA}{\textbf{0.7843}} & 	\cellcolor[HTML]{E6E6FA}{\textbf{0.0833}} & 	\cellcolor[HTML]{E6E6FA}{\textbf{0.0548}} & 	\cellcolor[HTML]{E6E6FA}{\textbf{8.39}} & 	\cellcolor[HTML]{E6E6FA}{54.73M} \\ \bottomrule 
	\end{tabular}
\end{table}

\begin{table}[]
	\vspace{-1em}
	\centering
\caption{ Quantitative Comparison of Time Series Forecasting on GreenEarthNet. The best and second-best values are bolded and underlined respectively.}
\label{tab:predearthx}
\setlength\tabcolsep{1pt}
\scalebox{0.7}{
	\begin{tabular}{c|c|c|ccccc|c}
		\toprule
		\multirow{2}{*}{Methods} & \multirow{2}{*}{Conditions} & \multirow{2}{*}{Time series} & \multicolumn{5}{c|}{RGBN/NDVI} & \multirow{2}{*}{Params} \\
		&  &  & PSNR↑ & SSIM↑ & RMSE↓ & MAE↓ & SAM↓ &  \\		\midrule
		\multicolumn{9}{c}{\cellcolor[HTML]{E7E7E7}\textit{\textbf{Non-generative model}}}                                                                                                                                                                                                                                    \\ \midrule
		PredRNN & His. RGBN & $\checkmark$  & -/18.57 & -/0.6557 & -/0.1263 & -/0.0919 & - & 1.4M \\
		SimVP & His. RGBN & $\checkmark$  & -/18.97 & -/0.6839 & -/0.1202 & -/0.0902 & - & 6.6M \\
		Contextformer & His. RGBN & $\checkmark$  & -/\underline{19.95} & -/\textbf{0.7058} & -/\underline{0.1091} & -/\underline{0.0791} & - & 6.1M \\
		STAU & His. RGBN & $\checkmark$  & 24.93/16.97 & 0.7245/0.6581 & 0.0598/0.1799 & 0.0408/0.0954 & 7.42/- & 7.63M \\
		\midrule
		\multicolumn{9}{c}{\cellcolor[HTML]{E7E7E7}\textit{\textbf{Generative model}}}                                                                                                                                                                                                                                    \\ \midrule
		Latte & His. RGBN & $\checkmark$  & \underline{27.23}/18.02 & \underline{0.8456}/\underline{0.6984} & \underline{0.0462}/0.1601 & \underline{0.0291}/0.0876 & \underline{4.59}/- & 32.32M \\
		LARP & His. RGBN & $\checkmark$  & 25.65/16.87 & 0.8238/0.6610 & 0.0551/0.1811 & 0.0339/0.0981 & 5.19/- & 58.56M \\
		DFoT & His. RGBN & $\checkmark$  & 23.34/15.71 & 0.7796/0.6340 & 0.0706/0.2081 & 0.0456/0.1293 & 6.56/- & 27.97M \\
		FAR & His. RGBN & $\checkmark$  & 24.29/17.53 & 0.8072/0.6814 & 0.0623/0.1773 & 0.0387/0.0914 & 5.37/- & 32.88M \\
		\cellcolor[HTML]{E6E6FA}{UniTS} & \cellcolor[HTML]{E6E6FA}{His. RGBN} & \cellcolor[HTML]{E6E6FA}{$\checkmark$}  & \cellcolor[HTML]{E6E6FA}{\textbf{31.14}/\textbf{20.13}} & \cellcolor[HTML]{E6E6FA}{\textbf{0.8667}/0.7006} & \cellcolor[HTML]{E6E6FA}{\textbf{0.0291}/\textbf{0.1071}} & \cellcolor[HTML]{E6E6FA}{\textbf{0.0181}/\textbf{0.0704}} & \cellcolor[HTML]{E6E6FA}{\textbf{4.43}/-} & \cellcolor[HTML]{E6E6FA}{54.73M} \\ \bottomrule 
	\end{tabular}}
\vspace{-2em}
\end{table}


\textbf{Quantitative Comparison.} We summarize the time series forecasting results of all methods in Tables~\ref{tab:pred}-\ref{tab:predearthx}. For the TS-S12 dataset, only 4 historical images are provided to predict future images with a sequence length ranging from 4$\sim$93. UniTS achieves the best performance overall, outperforming SyncVP by 1.52 dB in PSNR and reducing SAM by 1.54. On GreenEarthNet, UniTS obtains the best results in forecasting raw four-band Sentinel-2 reflectance. Compared to the discriminative model Contextformer, which directly predicts NDVI, the NDVI derived from UniTS does not achieve the best result in SSIM. This is primarily because directly learning the single-channel NDVI is a simpler regression task, whereas UniTS jointly models the more challenging distribution of raw reflectance across multiple bands.

\subsection{Ablation Study}

\begin{table}[]
	\vspace{-2em}
	\centering
	\caption{Ablation Comparison of Different Modules in UniTS.}
	\label{tab:ablation}
	\begin{tabular}{|ccc|ccc}
		\toprule
		\multicolumn{3}{c|}{Ablation Setting}               & \multicolumn{3}{c}{Time Series Cloud Removal/Forecasting} \\  \midrule
		\multicolumn{1}{c|}{Acor} & \multicolumn{1}{c|}{STM} & Metadata & PSNR↑       & SSIM↑         & SAM↓       \\ \midrule
		\multicolumn{1}{c|}{}     & \multicolumn{1}{c|}{}    &          & 17.58/19.17 & 0.6910/0.7165 & 9.77/12.68 \\
		\multicolumn{1}{c|}{$\checkmark$}    & \multicolumn{1}{c|}{}    &          & 19.02/20.45 & 0.7301/0.7338 & 8.23/12.04 \\
		\multicolumn{1}{c|}{}     & \multicolumn{1}{c|}{$\checkmark$}   &          & 18.79/20.63 & 0.7287/0.7411 & 9.16/11.54 \\
		\multicolumn{1}{c|}{$\checkmark$}    & \multicolumn{1}{c|}{$\checkmark$}   &          & 19.66/21.74 & 0.7451/0.7767 & 7.63/8.97  \\
		\multicolumn{1}{c|}{$\checkmark$} & \multicolumn{1}{c|}{$\checkmark$} & $\checkmark$ & \textbf{20.29/22.57}       & \textbf{0.7592/0.7843}       & \textbf{7.42/8.39}    \\ \bottomrule  
	\end{tabular}
\end{table}

\begin{table}[]
	\centering
	\caption{Ablation Comparison of Different Conditional Fusion.}
	\label{tab:fuse}
	\begin{tabular}{c|ccc}
		\toprule
		\multirow{2}{*}{Condition Fusion} & \multicolumn{3}{c}{Time Series Cloud Removal/Forecasting}          \\ \cmidrule(l){2-4} 
		& PSNR↑       & SSIM↑         & SAM↓       \\ \midrule
		Concat          & 19.02/20.64 & 0.7270/0.7383 & 8.74/11.66 \\
		Cross-attention & 18.96/20.81 & 0.7322/0.7407 & 8.91/11.17 \\
		Acor                               & \textbf{20.29/22.57} & \textbf{0.7592/0.7843} & \textbf{7.42/8.39} \\ \bottomrule  
	\end{tabular}
\vspace{-2em}
\end{table}

\begin{table}[]
	\centering
	\caption{Ablation Comparison of Sampling Steps in Flow Matching.}
	\label{tab:step}
	\begin{tabular}{c|ccc}
		\toprule
		\multirow{2}{*}{Samples steps} & \multicolumn{3}{c}{Time Series Cloud Removal/Forecasting} \\ \cmidrule(l){2-4} 
		& PSNR↑       & SSIM↑         & SAM↓      \\ \midrule
		10 & 20.29/22.57 & 0.7592/0.7843 & 7.42/8.39 \\
		20 & 20.23/22.51 & 0.7589/0.7844 & 7.43/8.39 \\
		30 & 20.27/22.59 & 0.7590/0.7845 & 7.40/8.37 \\
		40 & 20.30/22.56 & 0.7594/0.7841 & 7.43/8.40 \\ \bottomrule  
	\end{tabular}
\vspace{-2em}
\end{table}

\textbf{Effect of Metadata, Acor and STM.} Table~\ref{tab:ablation} presents the ablation results of different modules in UniTS. The introduction of the ACor module brings significant performance improvement, particularly in the time series cloud removal task, where the PSNR increases by approximately 1.44 dB, demonstrating its effectiveness in dynamically fusing multimodal conditional information. The STM module further optimizes spatiotemporal dependency modeling, especially in forecasting tasks. The metadata provides essential spatiotemporal priors for the model, playing a notable role in enhancing forecasting accuracy. When all three modules are enabled simultaneously, the model achieves the optimal performance, indicating functional complementarity among the modules and their collective contributions to building comprehensive spatiotemporal understanding and generative capabilities.

\textbf{Condition Fusion Strategy.} We compare the impact of different conditional fusion strategies on the performance of UniTS in Table~\ref{tab:fuse}. Simple feature concatenation and cross-attention mechanisms perform similarly in both time series cloud removal and forecasting tasks, indicating limitations of traditional fusion approaches in modeling spatiotemporal features. In contrast, the Acor-based fusion strategy achieves the best performance, demonstrating that Acor can more effectively coordinate multimodal conditional information by dynamically generating affine transformation parameters.

\textbf{Flow Matching Samples Steps.} In Table~\ref{tab:step}, we evaluate the impact of the number of sampling steps in the flow matching on generation performance. As the number of sampling steps increases from 10 to 40, only minor fluctuations are observed in PSNR, SSIM, and SAM, with no clear monotonic upward or downward trend. This indicates that UniTS can achieve high-quality sequence generation with only a small number of sampling steps.

\section{Conclusions}
\label{sec:conclusions}
In this paper, a universal spatiotemporal generative framework UniTS is proposed based on the flow matching paradigm, which for the first time achieves unified modeling of multiple time series tasks in remote sensing, including time series reconstruction, cloud removal, semantic change detection, and forecasting. The core architecture of UniTS is built on a diffusion transformer integrated with spatiotemporal blocks, where we design an Adaptive Condition Injector (ACor) to dynamically embed multimodal conditional information, enabling high-quality condition-guided generation, and introduce a Spatiotemporal-aware Modulator (STM) to enhance the model's capability of capturing complex dependencies through explicit spatiotemporal priors. Additionally, we construct two high-quality multimodal time series datasets, TS-S12 and TS-S12CR, by collecting data from tens of thousands of globally distributed ROIs, serving as important benchmarks for evaluating the performance of time series cloud removal and forecasting tasks in real-world scenarios. Extensive experiments across multiple tasks demonstrate that UniTS significantly outperforms existing specific models and foundational models. UniTS not only provides a powerful universal framework but also presents a new paradigm for earth observation spatiotemporal analysis driven by generative models.

\bibliographystyle{IEEEtran}
\bibliography{bibfile_zyx}

\end{document}